\documentclass[10pt,journal,compsoc]{IEEEtran}
\usepackage{color}
\usepackage{graphicx}
\usepackage{authblk}
\usepackage{caption}
\usepackage{cite}

\usepackage{multirow}

\usepackage{amsmath}
\DeclareMathOperator*{\argmin}{arg\,min}
\newtheorem{definition}{Definition}

\usepackage{algorithm}
\usepackage[noend]{algpseudocode}

\ifCLASSOPTIONcompsoc
  \usepackage[caption=false,font=normalsize,labelfont=sf,textfont=sf]{subfig}
\else
  \usepackage[caption=false,font=footnotesize]{subfig}
\fi
\usepackage[bottom]{footmisc}

\begin{document}

\title{Privacy-Preserving Collaborative Learning through Feature Extraction}

\author{Alireza~Sarmadi,
        Hao~Fu,
        Prashanth~Krishnamurthy,~\IEEEmembership{Member,~IEEE,}
        Siddharth~Garg,~\IEEEmembership{Member,~IEEE,}
        and~Farshad~Khorrami,~\IEEEmembership{Senior~Member,~IEEE}
\thanks{The authors are with the Department of Electrical and Computer Engineering, NYU Tandon School of Engineering, Brooklyn NY, 11201. E-mail: \{as11986@nyu.edu, hf881@nyu.edu, prashanth.krishnamurthy@nyu.edu, sg175@nyu.edu, khorrami@nyu.edu\}}}

\IEEEtitleabstractindextext{%
\begin{abstract}
    We propose a framework in which multiple entities collaborate to build a machine learning model while preserving privacy of their data. The approach utilizes feature embeddings from shared/per-entity feature extractors transforming data into a feature space for cooperation between entities. We propose two specific methods and compare them with a baseline method. In Shared Feature Extractor (SFE) Learning, the entities use a shared feature extractor to compute feature embeddings of samples. In Locally Trained Feature Extractor (LTFE) Learning, each entity uses a separate feature extractor and models are trained using concatenated features from all entities. As a baseline, in Cooperatively Trained Feature Extractor (CTFE) Learning, the entities train models by sharing raw data. Secure multi-party algorithms are utilized to train models without revealing data or features in plain text. We investigate the trade-offs among SFE, LTFE, and CTFE in regard to  performance, privacy leakage (using an off-the-shelf membership inference attack), and computational cost. LTFE provides the most privacy, followed by SFE, and then CTFE. Computational cost is lowest for SFE and the relative speed of CTFE and LTFE depends on network architecture. CTFE and LTFE provide the best accuracy. We use MNIST, a synthetic dataset, and a credit card fraud detection dataset for evaluations.
\end{abstract}

}

\maketitle

\section{Introduction}
\label{sec.introduction}

Machine learning algorithms typically require large training datasets to have high accuracy on the test input dataset. Unfortunately, such large training datasets may not be always available to the trainer. One solution is to obtain training data from other entities. For example, two or more hospitals could share their training data with each other to collaboratively train a more accurate model than that trained by each individual hospital. However, hospitals may also want to preserve patients' privacy as much as possible. Therefore, there is a need for algorithms that enable data sharing without compromising privacy.

Federated Learning (FL) \cite{MMRHYB17, KJMYRSB16} is one approach that seeks to address this goal using a central server that coordinates training between multiple parties (or "local" nodes) with private training datasets. The central server maintains a global model that the local nodes use to perform local (and private) gradient updates with respect to their own training data. The central server then updates the global model by coalescing the local gradient updates. But the local gradient updates are a source of information leakage of the local nodes to the central server \cite{NSH19, song2020analyzing, wang2019beyond}.

An alternate approach is that each party encrypts its data before sending it to the central server, and all the computations are executed in the encrypted domain using encryption methodologies \cite{damgaard2012multiparty, gordon2015constant, rabin1989verifiable, pass2017formal, chillotti2017faster, mukherjee2016two, li2017multi, wu2020efficient, li2020npmml, bohli2013security}. The main bottleneck of encryption-based algorithms is the computation cost due to the communication overhead, which consists of several rounds of interactions. These algorithms become computationally expensive when the dataset is large, the network has huge trainable parameters, or the number of parties involved in training is large.

In this paper, we propose a new solution that seeks to address the computational bottleneck of privacy-preserving collaborative learning using feature extraction. The methods we propose are Shared Feature Extractor Learning (SFE) and Locally Trained Feature Extractor Learning (LTFE) along with the baseline solution that we refer to as Cooperatively Trained Feature Extractor Learning (CTFE). Our implementations of these approaches are based on Secure Multi-Party Computation (SMPC). SMPC enables the parties to collaboratively perform computations without revealing raw data and therefore ensures input privacy, i.e., protects against direct leakage of raw training data by enabling computations over secretly shared data rather than raw data. 

\begin{figure*}
	\centering
    \includegraphics[width=0.95\linewidth]{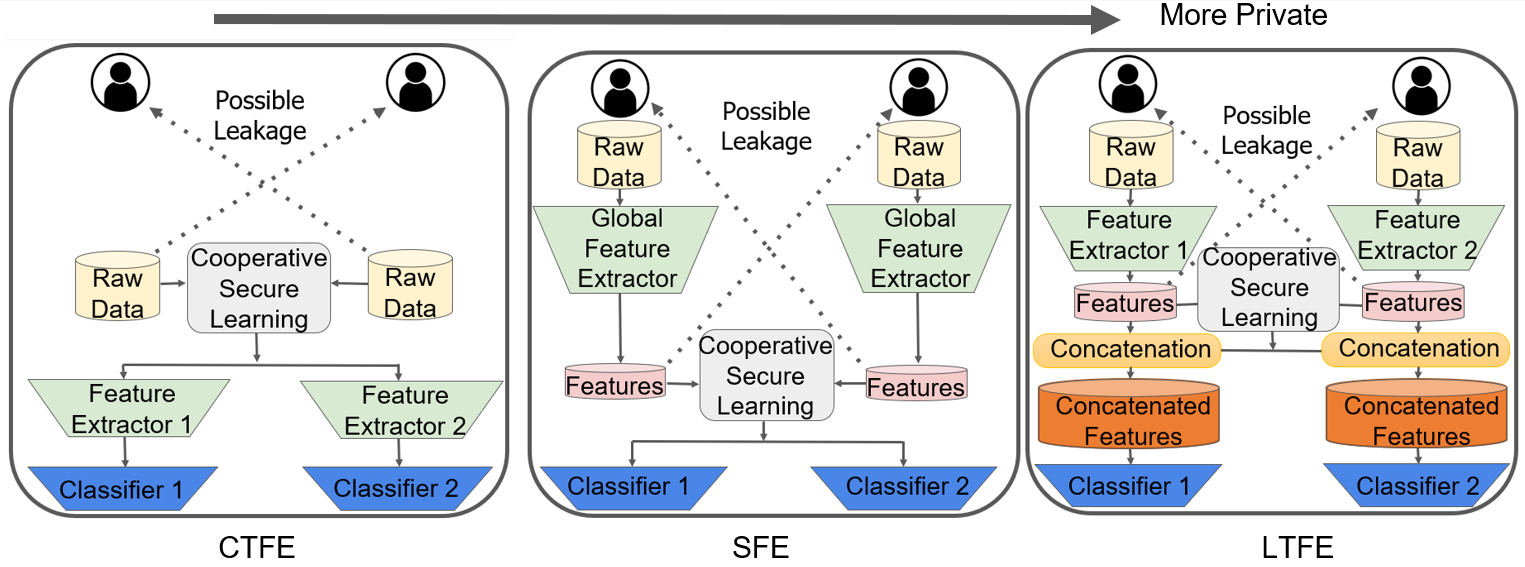}
	\caption{Cooperative secure learning scenarios. Left: CTFE. Middle: SFE. Right: LTFE.}
	\label{fig:sfeltfe}
\end{figure*}

CTFE is a baseline method in which each party encrypts (using additive secret sharing) a portion of its local data samples in the input space and shares the encrypted data with other parties. Then, other parties train their models with their local data and the shared encrypted data. The proposed SFE and LTFE methods compared to CTFE are shown in Fig.~\ref{fig:sfeltfe}. SFE reduces the computation costs of CTFE by training the feature extractor publicly (on training data available to all parties) and training only the classifier in the secure domain. Thus, cryptographic costs only incur in the classifier layers of the network. In LTFE, the goal is to provide greater privacy compared to CTFE. Here, each party trains its own local feature extractor with its own local private data. Then, each party trains its own classifier by concatenating encrypted features from all parties. In LTFE, each party also performs secure inference using its own locally computed features and encrypted features from other parties. The privacy benefits of LTFE over CTFE arise from the fact that only the classifier stage is collaboratively trained, but the feature extractors are kept private. We focus on horizontal data partitioning (i.e., parties have different sets of samples) in the development of the proposed methods. However,  LTFE can also be applied to vertically partitioned data (i.e., each party has access to a subset of the features for the same set of samples shared among parties) since its architecture is based on concatenating outputs of feature extractors from the different entities.

\subsection{Our Contributions}
The contributions of this paper include: 
\begin{itemize}
    \item Two collaborative learning methods based on feature extraction (SFE, LTFE) are proposed and compared with a baseline method (i.e., CTFE).
    \item The proposed methods address the computational bottleneck of privacy-preserving collaborative learning using feature extraction.
    \item The LTFE differs from any existing methods by incorporating an extra feature-concatenation layer that increases the training security and model accuracy (without even sharing any data since each party benefits from the other parties' embeddings).
    \item These methods are evaluated on three datasets.
    \item Challenges, such as highly-unbalanced training data, accelerating training time, and increasing network accuracy are addressed.
    \item The privacy aspects of these algorithms are evaluated by utilizing an off-the-shelf attack model.
\end{itemize}



\section{Related Work}
\label{sec.related}

Several methods have been proposed in order to allow entities to share privately their training data for collaborative learning. In \cite{schiff2017screening}, the authors proposed a serial training methodology. However, this algorithm suffers from catastrophic forgetting. A second approach, proposed in \cite{gupta2018distributed,vepakomma2018split}, is split learning. The main idea is that during the forward pass, each party computes the output up to a layer called the cut layer and passes that to other parties or a server. Then during the backward pass, the gradients are calculated till the cut layer and after that, the gradients will be backpropagated on each party's side. Federated learning (FL) \cite{MMRHYB17, KJMYRSB16, ZLLSCC18, BEGHIIKKMM19, MSS19, bvhes20} is another distributed learning algorithm that aims to preserve training data privacy. 
However, it has been shown that client information can be reconstructed by a malicious server \cite{song2020analyzing, NSH19}.

Another approach, referred to as Privacy Preserving Data Publishing (PPDP) \cite{fung2010privacy}, sanitizes (or anonymizes) data before passing it to other parties in a way that the anonymized data itself does not reveal sensitive information of individuals. One well-known model is the \textit{k}-anonymity model of privacy \cite{samarati1998protecting} which led to other privacy models such as l-diversity \cite{machanavajjhala2007diversity}, and t-closeness \cite{li2007t}. The \textit{k}-anonymity model requires that at least $k$ samples in the anonymized dataset have the same feature values. Intuitively, this model makes sure that each individual is indistinguishable from at least k-1 individuals. However, this method has limitations such as difficulties in choosing sensitive attributes, curse of dimensionality, and susceptibility to attribute disclosure. PPDP algorithms achieve privacy but come at the expense of reduced utility, for example, lower model accuracy \cite{aggarwal2008general}. On the other hand, Privacy Preserving Data Mining (PPDM) \cite{ASR15} algorithms consider the knowledge about the mining algorithm to increase the utility of the data while achieving a level of data privacy. These algorithms deal with preserving privacy in the output of queries about the data without revealing any information about the actual data samples, while in PPDP this preservation happens in the input domain before sharing the data with the public. One of the well-studied PPDM methods is Differential Privacy (DP) \cite{dwork2006differential} in which the author proposed DP as a measure of information leakage between two sets of data that differ in only one specific sample. To achieve these goals, DP methods add noise from a carefully chosen distribution to dataset entries. Several works \cite{ACGMMTZ16, TF20, ZHH20, DDR20, zhao2019privacy, gursoy2019secure} have proposed a machine learning framework for training models with differential privacy. DP-based machine learning methods deal with information leakage from a trained model and do not provide information leakage guarantee between collaborating parties.

An orthogonal approach, which we adopt in this paper, is to leverage methods from cryptography; in particular, Homomorphic Encryption \cite{gentry2009fully}, and SMPC \cite{yao1982protocols}. In Homomorphic Encryption, a client encrypts and sends its data to a central server; the server computes directly on encrypted data and sends the results back to the client for decryption \cite{gentry2009fully, graepel2012ml, hesamifard2017cryptodl, damgaard2012multiparty, chillotti2017faster, yu2013toward}. Thus the server learns nothing about the client's input. In SMPC, two or more parties want to compute a joint function of their private inputs and reveal only the output of the function. For instance, SPDZ \cite{damgaard2012multiparty}, SecureNN \cite{wagh2019securenn}, SecureML \cite{mohassel2017secureml}, GAZELLE \cite{juvekar2018gazelle}, and MiniONN \cite{liu2017oblivious} are well known protocols in this area. SMPC has been utilized in various applications such as feature-based spam detection \cite{RRDNA22} using classification models owned by a third-party and cloud-based computation of pixel-wise features of private images \cite{XMSSXJ18}. Additionally, there have been efforts to reduce the computational cost of SMPC, such as using high fan-in gates for Garbled Circuits \cite{ball2016garbling}, lowering bandwidth using function sharing \cite{boyle2018limits}, and using mixed protocols \cite{mohassel2018aby3}. \cite{imani2019framework} exploits high-dimensional computing for encoding data and uses SMPC to share global keys between parties instead of securely sharing the original data. In this work, we exploit the SMPC setting for sharing data between parties, particularly, we consider the SPDZ protocol which is suitable for training neural networks.


\section{Problem Statement}
\label{sec.formulation}
We consider a setting with $p$ parties, each of which has a set of labeled data called $D_i=\{(x_{ij},y_{ij})\}_{j=1}^{N_i}$, where $i=1,...,p$, $N_i$ is the size of $i^{th}$ party's dataset, $x_{ij}$ is the $j^{th}$ sample, and $y_{ij}$ is its corresponding label. Each party seeks to train a classifier $g_i^{*}: R^{Q} \rightarrow \{0,1\}^K$ where $Q$ is the dimension of the inputs, and $K$ is the number of classes. Each party minimizes empirical risk to train its classifier,
\begin{equation}
g_i^{*}=\argmin_{g \in G} \frac{1}{N_i} \sum_{j=1}^{N_i}L(g(x_{ij}),y_{ij})
\label{eq:main_problem}
\end{equation}
where $L(.,.)$ is the loss function, and $G$ is a set of parameterized classifiers. \textit{The problem we seek to answer is how parties can improve their classification accuracy by sharing their training datasets (or features derived from the datasets) with each other, but without compromising privacy.} The proposed methods along with the baseline method are described in Sec. \ref{sec.method}.

\noindent
\textbf{Threat Model}: The threat model considered is as follows:
\begin{itemize}
    \item {\em Parties:} We consider a semi-honest (honest-but-curious) adversary model, in which all the parties (including the adversary party) follow the protocol strictly during training, and none of the parties colludes with each other. 
    The collaborative learning is based on an SMPC approach in which the privacy of the inputs is protected. 
    \item {\em Adversary:} After collaborative learning, one of the parties is considered to be the adversary that attempts to extract information on training data from the trained model. The adversary performs a membership inference attack on the trained model to try to identify samples that were present in the training data. The adversary does not have access to the model or its gradients during training.
    \item {\em Cryptography Server Provider (CSP):} The CSP used in SMPC (Sec.~\ref{sec:SMPC}) is  assumed to be trusted by all parties and does not collude with any party or perform any attacks \cite{ryffel2018generic, kamara2014scaling, carter2016secure}.
\end{itemize}


 

\noindent\textbf{Goals}: Our first goal is to design a methodology where all parties can collaborate with each other to improve their classification accuracy while keeping their data private. We will assume, however, that the classifier architecture and hyper-parameters are known to all the collaborating parties (e.g., if a neural network is going to be used, the number of layers and their dimensions will be known). Although SMPC offers a way to meet this goal, it can be extremely time-consuming; therefore, we seek to improve the computational cost of SMPC  solutions for collaborative training, but without compromising privacy or accuracy. The second goal is to evaluate the effectiveness of the proposed methodology in preventing information leakage of private training data with access to the output and the trained model by an honest-but-curious party (note that while SMPC  preserves the secrecy of shared data during training, it does not itself prevent information leakage from trained models).


\section{Secure Multi-party Computation (SMPC)}
\label{sec:SMPC}
SMPC enables a set of parties $P=\{1,\hdots,p\}$ to calculate a function $F(x_1, \cdots, x_p)=y$ such that each party only knows $y$ at the end of computation and $x_i$ (i.e., the $i^{th}$ party's secret input) is unknown to other parties. There are different ways to implement SMPC, such as garbled circuit \cite{yao1986generate}, and secret sharing \cite{shamir1979share}.
Moreover, to overcome the computational cost of having more than two parties, we have considered a server-aided setting \cite{kamara2011outsourcing} with a non-colluding server. In the server-aided setting, the server is referred to as cryptography server provider \cite{nikolaenko2013privacy}.

To intuitively understand the security guarantees of SMPC, consider a two-party setting where $x_1$ and $x_2$ are inputs from parties 1 and 2, respectively. The protocol preserves the privacy if each party learns nothing more than $f(x_1, x_2)$ assuming that neither party deviates from the protocol. This is called the honest-but-curious model. There are many methods for designing secure protocols as mentioned in Sec.~\ref{sec:SMPC}. In this paper, we use SPDZ protocol which exploits secret sharing and guarantees the following security definition \cite{evans2017pragmatic}:

\begin{definition}
The protocol $\Pi$ securely computes $f$ in honest-but-curious model if it satisfies the following two security guarantees:
\begin{enumerate}
    \item Correctness: The output of the protocol is correct if for any inputs $x_1$, $\cdots$ and $x_p$ the output of the protocol (i.e., $\Pi^f(x_1, \cdots ,x_p)$) is same as $f(x_1,\cdots,x_p)$.
    \item Security: We require that an honest-but-curious party does not learn anything about other parties' data. Therefore, we require the existence of an efficient simulator $Sim_i$ such that $View_i^{\Pi} \approx Sim_i(x_i, f(x_1, \cdots ,x_p))$ where $View_i^{\Pi}$ represents the view of the client in the execution of $\Pi$, and $\approx$ denotes computationally indistinguishable. \footnote{We use the formal definition of security \cite{evans2017pragmatic} where $Sim$ denotes a simulator algorithm in which the simulator is an ideal-world adversary. 
    The simulator's existence proves that the adversary can not accomplish anything in the real world that could not also be accomplished in the ideal world.
    }
\end{enumerate}
\end{definition}


In addition to the $p$ parties (in this paper $p=2$), we utilize an additional CSP in our implementations.
This CSP setting is used in our work since we implement SMPC using PySyft \cite{ryffel2018generic}, in which a CSP is used for the consistent generation of random numbers. 

\section{Methods for Collaborative Learning}
\label{sec.method}
To overcome the challenges mentioned in Sec. \ref{sec:SMPC}, the parties use a mapping from input space to a feature space (the mapping is called feature extractor) $f_i: R^{Q} \rightarrow R^{q}$ where $q$ is the dimension of feature space. Typically $q < Q$ which results in a feature extractor that transforms the input samples into a lower dimension space, thus, the mapping is not invertible. Then the features are used to find the best possible classifier $h_i: R^{q} \rightarrow \{0,1\}^K$. Therefore, $g_i = h_i \circ f_i$. We propose three scenarios in which the feature extractor and classifier are trained. These scenarios/methods provide different trade-offs between performance, speed, and privacy. 
\subsection{CTFE Learning}
The first scenario is that each party trains the feature extractor and the classifier simultaneously as
\begin{equation}
    \begin{array}{r}
    g_i^{*}=\argmin_{g \in G} \frac{1}{N_i} \sum_{j=1}^{N_i}L(g(x_{ij}),y_{ij}) + \\
    \frac{1}{N_{other}} \sum_{u=1}^{N_{other}}L(g(x_{vu}),y_{vu})
    \label{eq:ctfe}
    \end{array}
\end{equation}
where $(x_{vu}, y_{vu})$ denote samples from the set
\begin{equation}
    D_{other}=\{ (x_{vu}, y_{vu}) | (x_{vu}, y_{vu}) \in D_v^{'} \}
    \label{eq:other_set}
\end{equation}
where $v \neq i$, $D_v^{'} \subset D_v$, and $N_{other}$ is the size of this $D_{other}$. $D_v'$ denotes the portion of $D_v$ that is contributed by party $v$ for collaborative learning of other parties. We consider CTFE as a baseline scenario for the problem formulation in Sec. \ref{sec.formulation}. In CTFE, the parties exchange the raw data using SMPC.

\subsection{SFE Learning}
In this method, the parties come to an agreement to use a set of data available to all parties called $D_{shared}$ to find a general feature extractor as 
\begin{equation}
    g_{sh}^{*}=\argmin_{g \in G} \frac{1}{N_{sh}} \sum_{j=1}^{N_{sh}}L(g(x_{sh j}),y_{sh j})
    \label{eq:sfe_feature_ex}
\end{equation}
where $(x_{sh j}, y_{sh j})$ is the $j^{th}$ sample in $D_{shared}$, $N_{sh}$ is the size of $D_{shared}$, and the feature extractor of $g_{sh}^{*}$ (i.e., $f_{sh}^{*}$) is considered as the global feature extractor available to all the parties. It should be noted that such an agreement is  common in collaborative learning settings \cite{yang2019federated, DTSKHR18}. 
Each party can choose what kind of data they want to share. While the utility of cooperative learning is most useful when the data from other parties is different from its own data distribution, parties can choose to hide data they consider critical from other parties. In this context, it is to be noted that in many applications, anomalous data is less confidential since they are non-legitimate data introduced by an adversary.
 Then, the parties train their classifiers cooperatively by exchanging feature embeddings (using SMPC) as
\begin{equation}
    \begin{array}{r}
    h_i^{*}=\argmin_{h \in H} \frac{1}{N_i} \sum_{j=1}^{N_i}L(h(t_{ij}),y_{ij}) + \\
    \frac{1}{N_{other}} \sum_{u=1}^{N_{other}}L(h(t_{vu}),y_{vu})
    \label{eq:sfe_feature_class}
    \end{array}
\end{equation}
where $H$ is the set of candidate classifiers, $t_{ij}=f_{sh}^{*}(x_{ij})$, and $t_{vu}=f_{sh}^{*}(x_{vu})$ where $(x_{vu}, y_{vu})$ is as in \eqref{eq:other_set}.

\subsection{LTFE Learning}
In this method, unlike in SFE where the global feature extractor is shared and trained using data available to all parties, each party trains their own feature extractor using their data as
\begin{equation}
    g_i^{*}=\argmin_{g \in G} \frac{1}{N_{i}} \sum_{j=1}^{N_{i}}L(g(x_{ij}),y_{ij})
    \label{eq:ltfe_feature_ex}
\end{equation}
where the feature extractor of $g_i^{*}$ (i.e., $f_i^{*}$) is considered as the $i^{th}$ party's feature extractor. Since the feature extractors in LTFE are separately trained, the feature embeddings from different parties' feature extractors are not directly compatible/aligned between the parties. Hence, to be able to leverage the knowledge built into other parties' feature extractors, the feature embeddings computed by the parties' feature extractors are concatenated (as shown in Fig.~\ref{fig:sfeltfe}) and the overall concatenated feature vector forms the input to the classifier in LTFE. For example, consider party 1's first data sample (i.e., $x_{11}$). The sample will be passed to each feature extractor to get the corresponding feature embeddings (i.e., $fe_1=f_1^{*}(x_{11})$ and $fe_2=f_2^{*}(x_{11})$). Then these feature embeddings are concatenated to form a vector $w_{11}=[fe_1;fe_2]$. Therefore, the parties make an agreement to design their classifier which has the input size of $2 \times q$ and each party trains their own classifier as
\begin{equation}
    \begin{array}{r}
    h_i^{*}=\argmin_{h \in H} \frac{1}{N_i} \sum_{j=1}^{N_i}L(h(w_{ij}),y_{ij}) + \\
    \frac{1}{N_{other}} \sum_{u=1}^{N_{other}}L(h(w_{vu}),y_{vu})
    \label{eq:ltfe_class}
    \end{array}
\end{equation}
where $w_{ij} = [f_1^{*}(x_{ij}), f_2^{*}(x_{ij})]$ is a concatenation of feature embeddings from each party's feature extractor (considering 2 parties for simplicity), and $w_{vu} = [f_1^{*}(x_{vu}), f_2^{*}(x_{vu})]$ where $(x_{vu}, y_{vu})$ is as in \eqref{eq:other_set}. The parties exchange the concatenation of feature embeddings using SMPC.


\subsection{Neural Networks as Candidate Functions}
In this paper, neural networks are considered as candidate functions. The best candidate could be found by minimizing the empirical risk on the data as
\begin{equation}
    {\theta}_i^{*}=\argmin_{\theta_i} \frac{1}{N_i} \sum_{j=1}^{N_i}L(g(x_{ij};\theta_i),y_{ij})
\end{equation}
where ${\theta}_i$ is the vector of the neural network parameters (weights) corresponding to the $i^{th}$ party. For this purpose, we use Stochastic Gradient Descent (SGD). In Algorithms \ref{alg:ctfe}, \ref{alg:sfe}, and \ref{alg:ltfe}, the detailed procedure of CTFE, SFE, and LTFE are described, respectively. In all the algorithms, the weights' update of the models using the SGD is represented by functionality $GU$. The number of batches and the number of epochs are represented by $n_{batches}$, and $n_{epochs}$, respectively. 
In Algorithm \ref{alg:ctfe}, the $CTFE\_MAIN$ is the main procedure in which each party trains its model ($M_i$ corresponds to $i^{th}$ party's model) using its own data in non-secure mode and then continues training on the data from other parties in secure mode. In Algorithm \ref{alg:sfe}, in the main procedure ($SFE\_MAIN$), a general feature extractor ($M_{s}$) is trained using the shared data ($D_{s}$) available to all the parties. Then each party passes its own data through this feature extractor to find the feature embeddings ($F_{own}$ corresponds to the $i^{th}$ party feature embeddings, and $F_{other}$ shared by all other parties). Then each party trains its classifier using its own features and then continues training collaboratively the classifier ($M_i$ corresponds to $i^{th}$ party's classifier) using the feature embeddings from other parties in secure mode. In Algorithm \ref{alg:ltfe}, $LTFE\_MAIN$ is the main procedure. The parties first train their feature extractors ($N_i$ corresponds to the $i^{th}$ party feature extractor) on their local data. Then each party trains its classifier ($M_i$ corresponds to $i^{th}$ party model) on a concatenation of secure features from all the parties' feature extractors. This has been shown by the $Concatenate$ function that concatenates feature vectors from all parties horizontally for each sample. While the $Append$ function concatenates these vectors vertically. It should be noted that in all these algorithms, $\Pi$ denotes the implementation using SMPC, i.e., replacing $\Pi^{f}$ with $f$ throughout yields the non-secure (i.e., plain-text) implementations.

\begin{algorithm}
\caption{CTFE}
\label{alg:ctfe}
\begin{algorithmic}[1]
\Require{$D_1, D_2$}
\Ensure{$M_1, M_2$}

\Procedure{Train\_Model}{$M$, $D$}
   \For {$i \gets 1$ to $n_{epochs}$}
       \For {$j \gets 1$ to $n_{batches}$}
           \State $GU(M, D^j)$ 
           \Comment{Updating the weights of M using SGD}
       \EndFor
   \EndFor
   \State \Return $M$
\EndProcedure 
\Procedure{CTFE\_Main}{$D_1, D_2$}
\For {$i \gets 1$ to $2$}
    \State $M_i \leftarrow TRAIN\_MODEL(M_{i}, D_i)$
    \State $M_i \leftarrow \Pi^{TRAIN\_MODEL}(M_{i}, \{D_j^{'} |j\neq i, D_j^{'} \subset D_j\})$
\EndFor
\State \Return $M_1$, $M_2$
\EndProcedure
\end{algorithmic}
\end{algorithm}

\begin{algorithm}
\caption{SFE}
\label{alg:sfe}
\begin{algorithmic}[1]
\Require{$D_{s}$, $D_1, D_2$}
\Ensure{$M_{s}$, $M_1, M_2$}

\Procedure{Train\_Feature\_Extractor}{$M$, $D$}
    \For {$i \gets 1$ to $n_{epochs}$}
        \For {$j \gets 1$ to $n_{batches}$}
            \State $GU(M, D^j)$ 
            \Comment{Updating the weights of M using SGD}
        \EndFor
    \EndFor
    \State \Return $M$
\EndProcedure
\Procedure{Train\_Classifier}{$M$, $F$}
   \For {$i \gets 1$ to $n_{epochs}$}
       \For {$j \gets 1$ to $n_{batches}$}
           \State $GU(M, F^j)$ 
           \Comment{Updating the weights of M using SGD}
       \EndFor
   \EndFor
   \State \Return $M$
\EndProcedure 
\Procedure{SFE\_Main}{$D_{s}$, $D_1, D_2$}
\State $M_{s} \leftarrow$ TRAIN\_FEATURE\_EXTRACTOR($M_{s}$, $D_{s}$)
\For {$i \gets 1$ to $2$}
    \State $M_i \leftarrow TRAIN\_CLASSIFIER(M_{i},M_{s}(D_i))$
    \State $M_{i} \leftarrow \Pi^{TRAIN\_CLASSIFIER}(M_{i}, $ $ \{M_{s}(D_j^{'}) |j\neq i, D_j^{'} \subset D_j\})$
\EndFor
\State \Return $M_{s}$, $M_1$, $M_2$
\EndProcedure

\end{algorithmic}
\end{algorithm}

\begin{algorithm}
\caption{LTFE}
\label{alg:ltfe}
\begin{algorithmic}[1]
\Require{$D_1, D_2$}
\Ensure{$N_1, N_2$, $M_1, M_2$}

\Procedure{Train\_Feature\_Extractor}{$N$, $D$}
    \For {$i \gets 1$ to $n_{epochs}$}
        \For {$j \gets 1$ to $n_{batches}$}
            \State $GU(N, D^j)$ 
            \Comment{Updating the weights of N using SGD}
        \EndFor
    \EndFor
    \State \Return $N$
\EndProcedure

\Procedure{Train\_Classifier}{$M$, $F$}
   \For {$i \gets 1$ to $n_{epochs}$}
       \For {$j \gets 1$ to $n_{batches}$}
           \State $GU(M, F^j)$ 
           \Comment{Updating the weights of M using SGD}
       \EndFor
   \EndFor
   \State \Return $M$
\EndProcedure 

\Procedure{LTFE\_Main}{$D_1, D_2$}
\For {$i \gets 1$ to $2$}
     \State $N_i \leftarrow$ TRAIN\_FEATURE\_EXTRACTOR($N_i$, $D_i$)
\EndFor
\For {$i \gets 1$ to $2$}
    \State $F \gets 0$
    \Comment{Initialization of the array $F$}
    \For {$k \gets 1$ to $2$}
        \State $tmp \leftarrow 0$
        \Comment{Initialization of the array $tmp$}
        \For {$s \gets 1$ to $2$}
            \State  $tmp \leftarrow \Pi^{Concatenate}(tmp, \{N_{s}(D_k^{'}) | D_k^{'} \subset D_k\})$
        \EndFor
        \State $Append(F, tmp)$
    \EndFor
    \State $M_i \leftarrow \Pi^{TRAIN\_CLASSIFIER}(M_{i}, F)$
\EndFor
\State \Return $N_1$, $N_2$, $M_1$, $M_2$
\EndProcedure

\end{algorithmic}
\end{algorithm}

\subsection{Computational Complexity}
The proposed scenarios have different computational complexities. To have a fair comparison between them, we assume the feature extractors have the same architecture with 4 dense layers (including the input layer) and classifiers as well with 1 dense layer. Assume corresponding weights for each layer are as follows: $W_{n_1 Q}$, $W_{n_2 n_1}$, $W_{q n_2}$, and $W_{K q}$ where $n_i: i=1,2$ are the number of nodes in the first two layers after input layer in feature extractor. We train the model for $n$ epochs and we assume each party shares $t$ training samples. The time complexity of training such a neural network for each party is $\mathcal{O}(npt \times (Q n_1+n_1 n_2+n_2 q+ q K))$. In CTFE, each party trains the feature extractor and classifier together, therefore, the time complexity is of an order of $\mathcal{O}(npt \times (Q n_1+n_1 n_2+n_2 q+ q K))$. Training the model in SFE for each party is of an order of $\mathcal{O}(npt \times (q K))$ since the feature extractor is fixed during the classifier training. In LTFE, all the parties train their feature extractor simultaneously, however, for training the classifier, each party concatenates the feature embeddings from all the parties which makes the input to the classifiers higher in dimension (i.e., $p K$). Therefore, it is of order of $\mathcal{O}(npt \times (Q n_1+n_1 n_2+n_2 q+ q p K) + npt \times (p q K)) = \mathcal{O}(npt \times ( Q n_1+n_1 n_2+n_2 q+ (p+1)q K)$. Therefore, SFE has a lower computational burden compared to LTFE and CTFE. When the feature extractor is relatively simple (e.g., a few fully connected layers), LTFE is more computationally expensive than CTFE. For the case when the feature extractor has CNN layers, CTFE can be more computationally expensive than LTFE.

\subsection{Performance, Privacy, and Speed Trade-offs}
\begin{figure}
	\centering
	\includegraphics[width=0.75\linewidth]{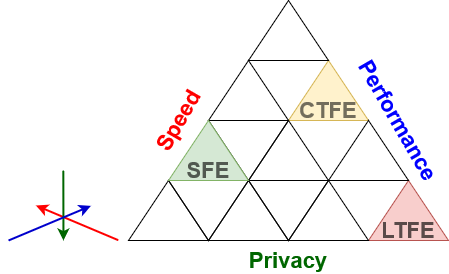}
	\caption{Trade-off between performance, privacy, and computational speed of the proposed scenarios. This figure shows the qualitative relationship between CTFE, SFE, and LTFE. The increase direction for each metric is depicted on the left side. It could be seen that LTFE is the most privacy-preserving, SFE and CTFE are faster than LTFE (the relative speed of CTFE and LTFE depends on the network architecture). In terms of performance, LTFE and CTFE perform better than SFE.}
	\label{fig:trade_off}
\end{figure}

LTFE has the highest performance (accuracy) under no or very little data sharing among the parties since during inference, each party benefits from the use of other parties' feature extractors. CTFE has the next higher performance and the last is SFE. However, as more data is shared among parties, the SFE approaches CTFE and both tend towards the performance of LTFE (especially if a large percentage of the data is shared). Lastly, the LTFE provides the highest level of privacy followed by SFE and last is the CTFE. The SFE has the lowest computational burden. LTFE is computationally more expensive than SFE since combining feature embeddings from feature extractors of different parties requires an additional step of concatenation, making the input to the classifiers higher in dimension. In Fig. \ref{fig:trade_off}, the trade-off among performance, privacy, and computational speed of the proposed scenarios is shown.

\section{Experimental Setup}
\label{sec.setup}
\subsection{Datasets}
We used three datasets to evaluate the scenarios described in Sec. \ref{sec.method}: MNIST and Credit Card Fraud Detection which are off-the-shelf datasets, and a synthetic dataset. Three different datasets are chosen to show the independence of the results from the choice of the dataset.

The MNIST dataset contains images of handwritten digits from 0 to 9. The dataset has 60,000 training samples and 10,000 test samples. The images are grayscale and of size $28 \times 28$.

The Credit Card Fraud Detection dataset contains anonymized credit card transactions labeled as fraudulent or genuine. The dataset has 284,807 samples that 492 of which are frauds. It also has 29 features that are obtained with PCA. 

The synthetic dataset is generated using the standard sklearn make\_classification module \footnote{sklearn.datasets.make\_classification}. The size of the feature vector is $784$, and the number of classes is 10. Then the samples are randomly split between parties. The synthetic dataset has 784 features and 10 classes.

For each dataset, we have split the dataset into 4 parts called global, party 1, party 2, and test data. For the synthetic dataset, each part has 20,000 samples. For the MNIST dataset, we split the 60,000 training samples into 12,600 samples for shared, and 23,700 samples for each party 1 and 2 as shown in Table \ref{table:size}. For the test data, we use the 10,000 test samples provided by the dataset. It should be noted that to see the benefit of sharing data between parties, we have split the training data in a way that party 1 has more samples for labels ZERO, ONE, and TWO, while it has fewer samples for labels SEVEN, EIGHT, and NINE. For party 2, it is vice-versa and shared data has more samples for labels THREE, FOUR, and FIVE than the other labels. For the fraud detection dataset, each party has 1,000 normal samples. The fraud samples are distributed as 100 for global data, 50 for party 1, 242 for party 2, and 100 for testing to study the benefit of sharing. The number of epochs is 10 for each of feature extractor training and the classifier training. We set each party to share 30\% of their data with other parties using the methods outlined in Sec.~\ref{sec.method}.

\begin{table}
\begin{center}
\caption{Data size for each party.}
\label{table:size} 
\begin{tabular}{cccccccc}
\hline\noalign{\smallskip}
Dataset & Global & Party 1  & Party 2 & Test  \\
\hline
MNIST & 12,600 & 23,700 & 23,700 & 10,000  \\
Synthetic & 20,000 & 20,000 & 20,000  & 20,000  \\
Fraud Detection & 1,100 & 1,050 & 1,242  & 1,100  \\
\hline
\end{tabular}
\end{center}
\end{table}


\subsection{Models}
A four-layer fully connected neural network (FCN) is used as the model to classify the pre-processed data. ReLU activation functions are used between each of the layers and for the last layer a semi-sigmoid activation function is used which has the following formula: $ReLU(x)-ReLU(x-1)$ where $x$ is the input vector to the layer. The semi-sigmoid activation function is used instead of sigmoid due to PySyft limitations. A neural network has three main components: the input layer, hidden layers, and classifier layers. The input layer and hidden layers together form the feature extractor. The single output layer functions as the classifier. The pre-processed data is passed through the three-layer feature extractor to generate the feature embeddings in both the SFE and LTFE. The layer sizes are 64, 64, 64, and 10, therefore, the dimension of the feature embeddings is 64. For all the experiments, Mean Squared Error (MSE) is used as the loss function, and SGD \cite{goodfellow2016deep} is used as the optimizer with a learning rate of 0.1, and an $L2$ regularization with a weight of 0.0002. For the MNIST dataset, we also evaluated the methods on the LeNet-5 \cite{lecun1998gradient} architecture in which the dimension of the feature embeddings is 120. However, due to the PySyft limitations, the proposed semi-sigmoid is used as the activation function for all the layers. It should be noted that to be consistent, the input images are zero-padded such that the inputs to the LeNet-5 have the dimension of $32 \time 32 \time 1$. The SGD with a learning rate of 0.1 is used for the optimizer.

\subsection{Performance Benefits of Cooperation}
\begin{figure*}
	\centering
	\subfloat[Scores for SEVEN]{\includegraphics[width=0.33\linewidth]{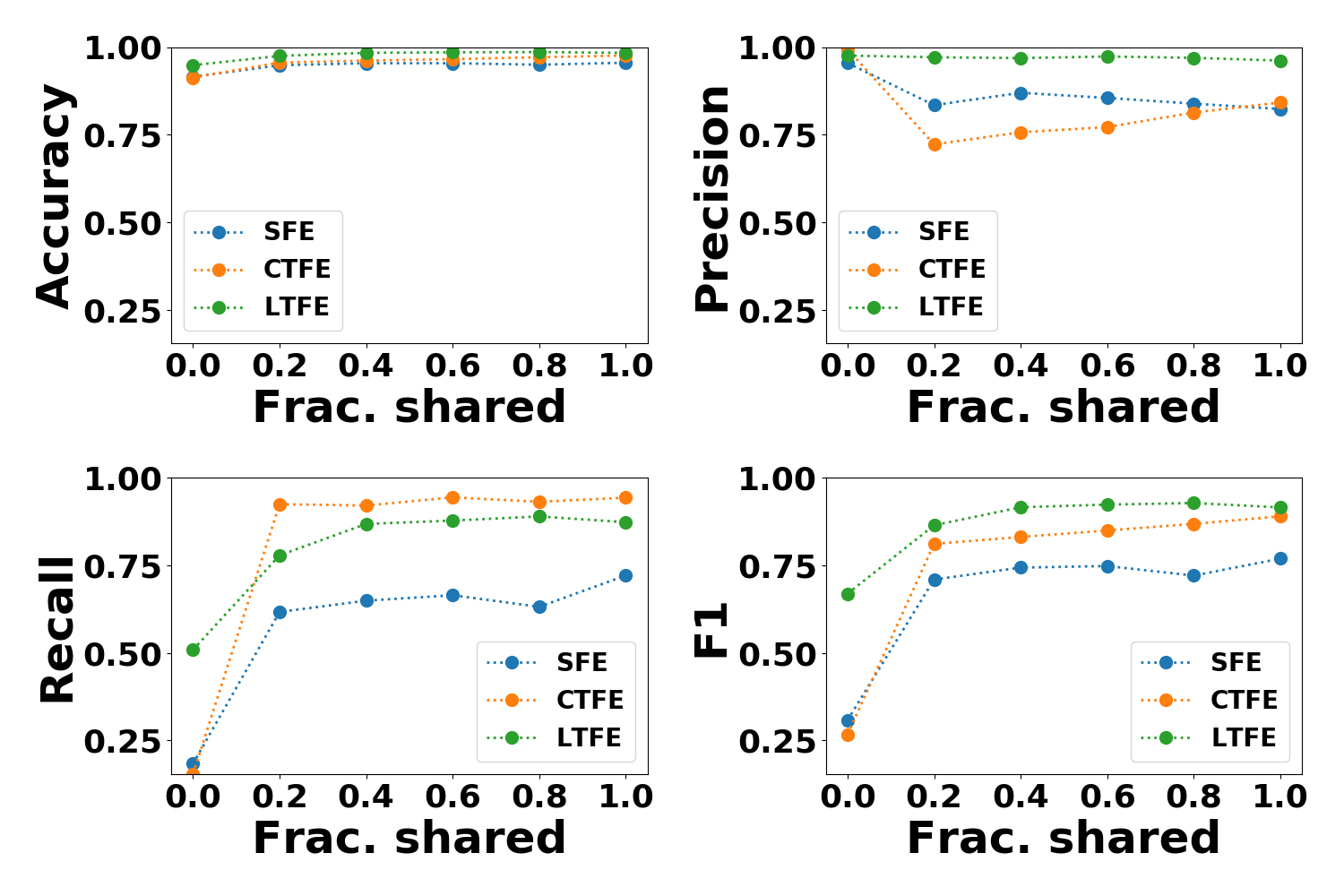}\label{a}}
	\hfill
	\subfloat[Scores for EIGHT]{\includegraphics[width=0.33\linewidth]{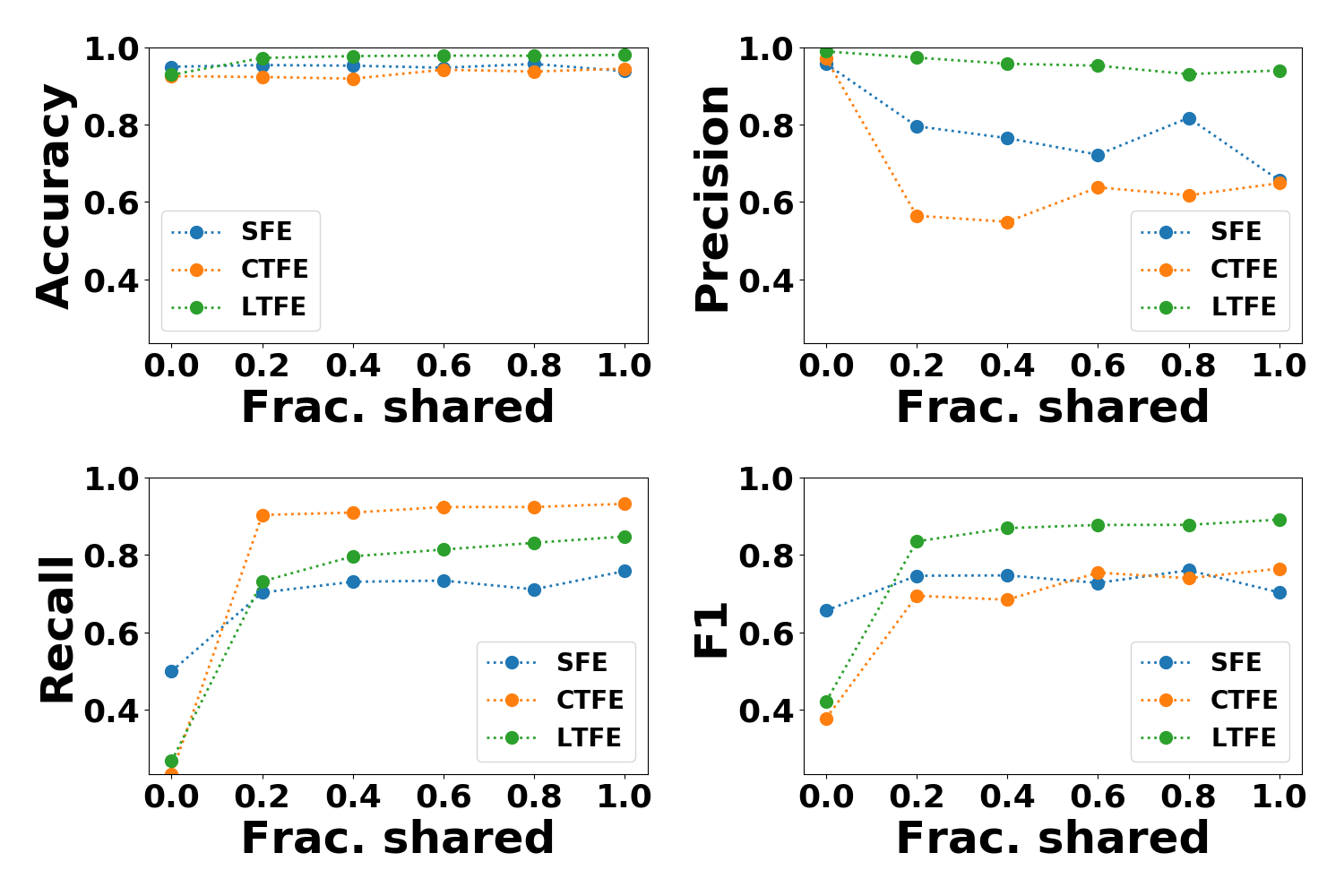}\label{b}}
	\hfill
	\subfloat[Scores for NINE]{\includegraphics[width=0.33\linewidth]{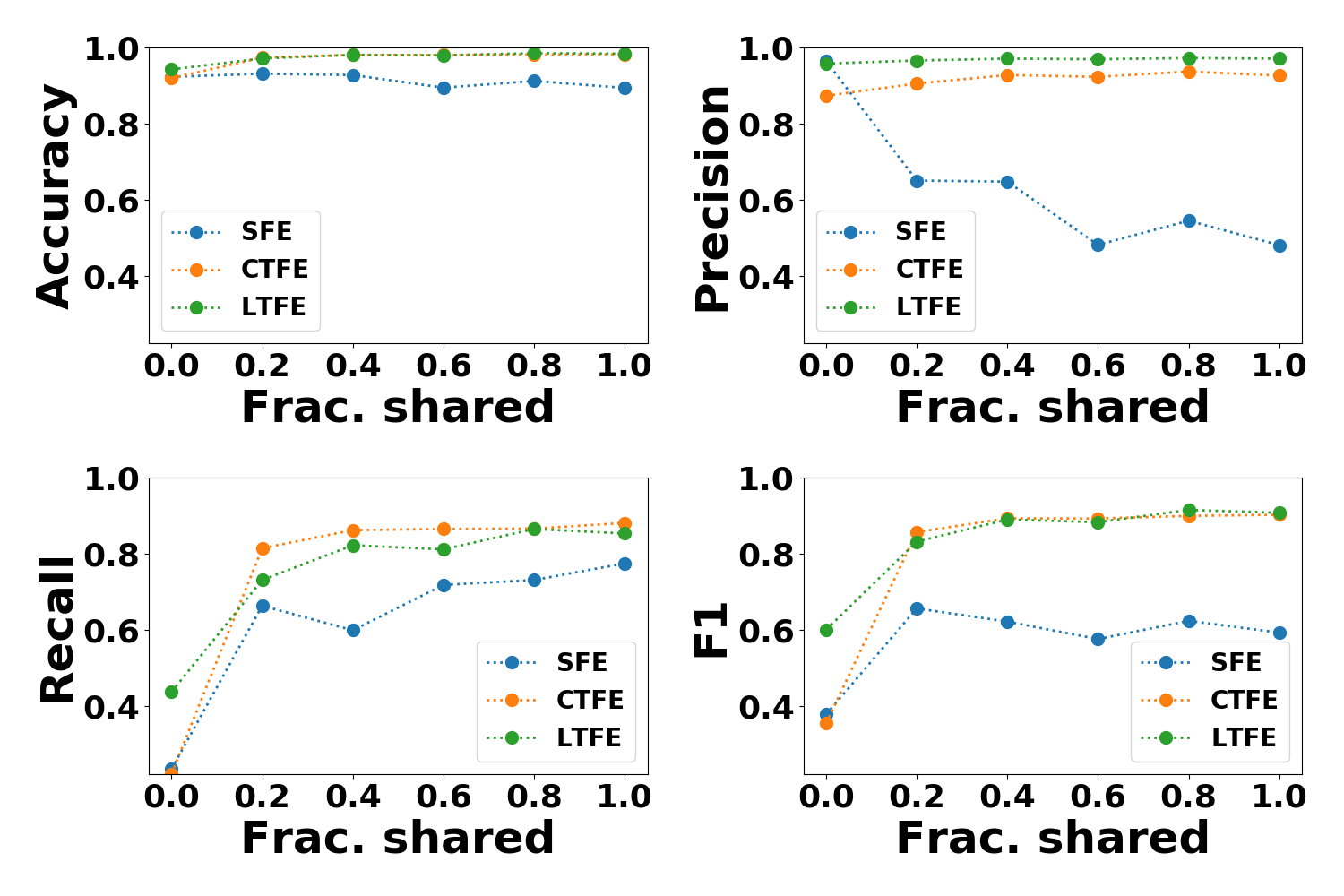}\label{c}}
	\caption{Performance with different sizes of MNIST shared dataset on the fully connected network for labels {\bf SEVEN}, {\bf EIGHT}, and {\bf NINE} in non-secure mode.}
	\label{fig:pct}
\end{figure*}

Each party benefits from the other parties' data in cooperative learning. The more samples are available from a data distribution, the more information is available for a machine learning algorithm to find the underlying distribution. Therefore, it is expected that feature embedding cooperation between parties has the same effect. To verify this, we have trained and evaluated the model when it has access to 0\% to 100\% of other party data with a step of 20\%. Fig.~\ref{fig:pct} shows the performance of party 1's trained model using CTFE, SFE, and LTFE on MNIST data with different percentages of data shared by party 2. The figure shows 4 performance metrics (i.e., accuracy, precision, recall, and F1 score) for samples with a specific label. It is seen that as more data are shared by other parties, the performance of the model increases. It should be noted that since most of the labels are common among all the parties, SEVEN, EIGHT and NINE are chosen in Fig.~\ref{fig:pct} which have different distributions among parties.

\subsection{Secure Computation Implementation}
In this work, we use PySyft \cite{ryffel2018generic} library which relies on SPDZ protocol \cite{damgaard2012multiparty}. SPDZ enables more complex operations than addition and multiplication, which makes it suitable for training neural networks. Also, we made the following changes to increase  training speed and model performance:
\begin{itemize}
    \item Modifying PySyft internals including modifying message compression settings to trade off computation vs. communication time and replacing multiprocessing-based parallelization (which we found to be a major cause of slowdowns) in SPDZ computations with multithreading (with a tuned optimal number of threads) which is more efficient under larger process memory footprints
    \item Increasing the \textit{precision\_fractional} variable. A large \textit{precision\_fractional} number allows the modified data to keep more information from the raw data. The original \textit{precision\_fractional} was 3 in PySyft. We changed it to 5. 
\end{itemize}

\subsection{Attack Models}
\label{subsec:attack_models}
\begin{figure*}
	\centering
	\subfloat[]{\includegraphics[width=0.33\linewidth]{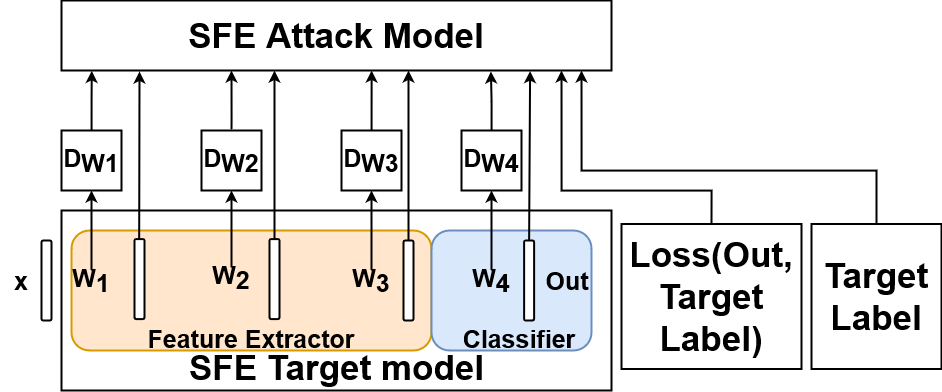}\label{a1}}
	\hfill
	\subfloat[]{\includegraphics[width=0.33\linewidth]{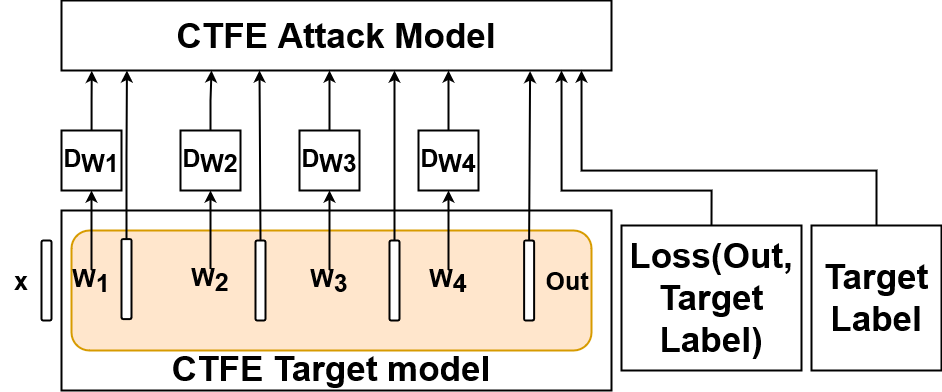}\label{b1}}
	\hfill
	\subfloat[]{\includegraphics[width=0.33\linewidth]{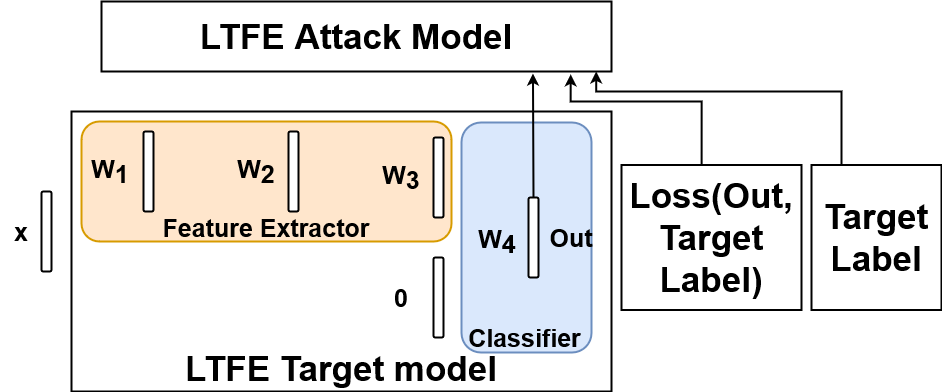}\label{c1}}
	\\
	\subfloat[]{\includegraphics[width=0.33\linewidth]{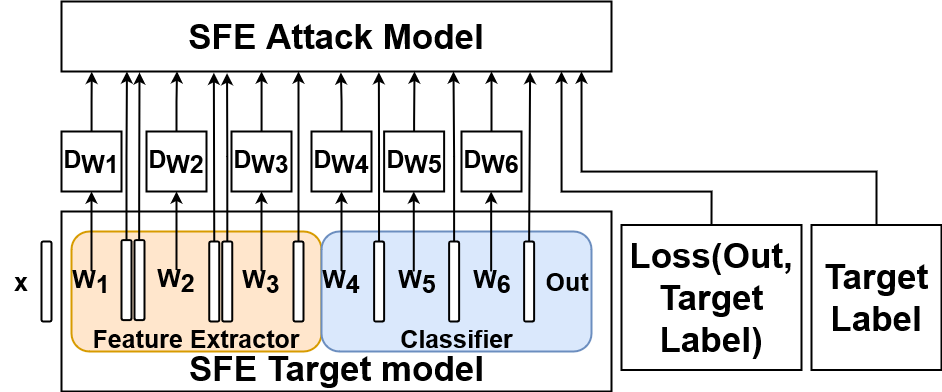}\label{a1_le}}
	\hfill
	\subfloat[]{\includegraphics[width=0.33\linewidth]{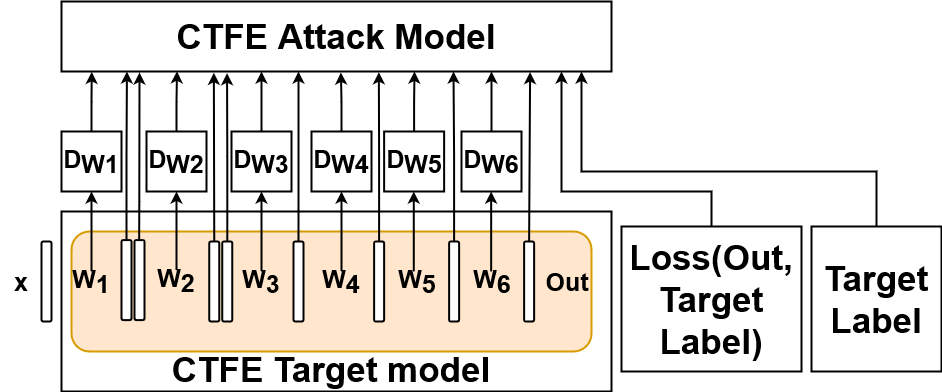}\label{b1_le}}
	\hfill
	\subfloat[]{\includegraphics[width=0.33\linewidth]{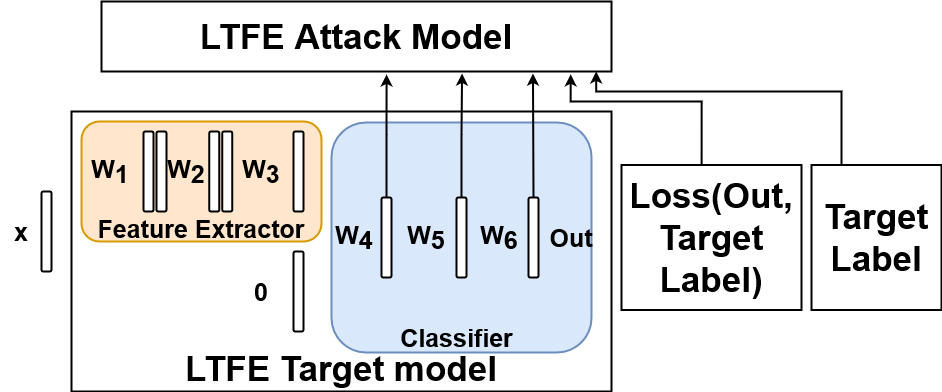}\label{c1_le}}
	\caption{Attack models are used for each model based on the availability of the model weights to the attacker. The top row corresponds to fully connected target model architecture, and the bottom row corresponds to LeNet-5 architecture in which each CNN layer and its subsequent average pooling layer are depicted with double blocks. Also, the gradient of loss function w.r.t $W_i$ is depicted with $D_{W_i}$.}
	\label{fig:attack_models}
\end{figure*}

To quantify the information leakage of the proposed scenarios, we considered an off-the-shelf membership inference attack model as a metric. The purpose of the attack model is to evaluate the algorithms in terms of the source of the data. In other words, a party attempts to identify samples from other parties' data by analyzing the model output. Without loss of generality, assume we have 2 parties collaborating to train their models using CTFE, SFE, and LTFE, and assume that the attacker is on party 1's side. For simplicity and to more clearly illustrate the membership inference attack, we consider the scenario where party 1 does not have its own data and only trains its model using all the other party's data. After the target models are trained, the attacker applies a membership inference attack algorithm. In this paper, we use the attack model described in \cite{NSH19}. However, based on the attacker's access to the target model, the attack model differs between the scenarios. The attack model for each scenario is depicted in Fig. \ref{fig:attack_models}. It could be seen that in CTFE, the attacker has access to the target model weights since the model is trained from input to output on the attacker's side. Also, in SFE, the attacker has access not only to the classifier but also to the shared model that is available to all parties. However, in LTFE, the attacker has access only to the classifier and does not have access to the other parties' feature extractors. For all the attack models, we use the same architecture and training procedure proposed in \cite{NSH19}.

\section{Experimental Results}
\label{sec.result}

In this section, we investigate the trade-offs between the three methods on the three datasets.

\subsection{Performance}
The performance of party 1's model using the proposed methods  and the Non-Cooperative (NC) case are shown in Tables~\ref{table:mnist_performance_fcn}, \ref{table:mnist_performance_lenet}, \ref{table:synth_performance}, and \ref{table:fraud_performance} for MNIST FCN, MNIST LeNet-5, synthetic dataset, and fraud detection cases, respectively. In these tables, the A, P, R, and F1 columns correspond to  accuracy, precision, recall, and F1 score, respectively. The NC in these tables corresponds to the case using CTFE where the percentage of sharing is 0\%  which means the model does not take advantage of other parties' data and only is trained on the local data. 
Comparison to the NC indicates improvements in model performance by using other parties' data. The LTFE  has an overall better performance than SFE. This is because LTFE uses concatenated features and therefore can benefit from the use of other parties' feature extractors, whereas SFE only uses a global feature extractor.

The training time and inference time are shown in Table~\ref{table:time} for all the settings. SFE takes the least time since it uses a shared feature extractor, which can be implemented in non-secure mode. Additionally, it can be seen that CTFE is more computationally expensive than SFE since, in CTFE, all of the model's weights are updated using gradient descent in secure mode while in SFE, only the classifier weights are updated. LTFE could be more computationally expensive than SFE because LTFE involves concatenating features that have to be implemented in secure mode, whereas other methods do not have such a process. The relative training time of CTFE and LTFE is dependent on the network architecture. It could be seen that for the MNIST dataset on the LeNet-5 network, CTFE is more computationally expensive since the feature extractor has CNN layers. However, when fully connected layers are used for the feature extractor, LTFE is more computationally expensive than CTFE.

\begin{table}
\begin{center}
\caption{Performance of party 1's models using SFE, CTFE, LTFE, and Non-Cooperative (NC) case on MNIST data (FCN network) with 30\% of sharing. A, P, R, and F1 correspond to accuracy, precision, recall, and F1 score, respectively. The results are shown for labels 7, 8, and 9 since party 1 has fewer samples for those labels than party 2, and for brevity.}
\label{table:mnist_performance_fcn} 
\begin{tabular}{c|cccccc}
\hline
Method & Label & A  & P & R & F1 \\
\hline
 \multirow{3}{*}{NC}
& SEVEN & 95.12 & 99.09 & 53.02 & 69.07\\
& EIGHT & 93.85 & 98.12 & 37.58 & 54.34\\
& NINE  & 93.60 & 95.33 & 38.45 & 54.80 \\
 \hline
  \multirow{3}{*}{SFE}
& SEVEN & 95.19 & 97.90 & 54.38 & 69.92 \\
& EIGHT & 94.40 & 97.05 & 43.84 & 60.40 \\
& NINE  & 94.35 & 96.25 & 45.79 & 62.06 \\
 \hline
\multirow{3}{*}{CTFE}
& SEVEN & 97.18 &  95.94 &  75.78 &  84.67 \\
& EIGHT & 96.10 &  83.56 &  74.64 &  78.85 \\
& NINE  & 96.18 &  91.97 &  68.09 &  78.25 \\
 \hline
\multirow{3}{*}{LTFE}
& SEVEN & 97.82 & 98.21 & 80.25 & 88.33 \\
& EIGHT & 97.70 & 95.59 & 80.08 & 87.15 \\
& NINE  & 97.69 & 97.91 & 78.79 & 87.31 \\
\hline
\end{tabular}
\end{center}
\end{table}

\begin{table}
\begin{center}
\caption{Performance of party 1's models using SFE, CTFE, LTFE, and Non-Cooperative (NC) case on MNIST data (LeNet-5 network) with 30\% of sharing. A, P, R, and F1 correspond to accuracy, precision, recall, and F1 score, respectively. The results are shown for labels 7, 8, and 9 since party 1 has less number of samples for those labels than party 2, and for brevity.}
\label{table:mnist_performance_lenet} 
\begin{tabular}{c|cccccc}
\hline
Method & Label & A  & P & R & F1 \\
\hline
 \multirow{3}{*}{NC}
& SEVEN & 97.89 & 99.04 & 80.25 & 88.66\\
& EIGHT & 98.15 & 99.13 & 81.72 & 89.59\\
& NINE  & 98.47 & 99.65 & 85.13 & 91.82\\
 \hline
  \multirow{3}{*}{SFE}
& SEVEN & 98.20 & 85.57 & 99.22 & 91.89 \\
& EIGHT & 97.94 & 83.33 & 98.56 & 90.31 \\
& NINE  & 99.12 & 95.37 & 95.94 & 95.65 \\
 \hline
\multirow{3}{*}{CTFE}
& SEVEN & 98.57 & 88.31 & 99.22 & 93.45 \\
& EIGHT & 98.67 & 88.90 & 98.67 & 93.53 \\
& NINE  & 99.28 & 95.18 & 97.82 & 96.48 \\
 \hline
\multirow{3}{*}{LTFE}
& SEVEN & 99.72 & 99.50 & 97.76 & 98.63 \\
& EIGHT & 99.80 & 99.48 & 98.46 & 98.97 \\
& NINE  & 99.64 & 99.80 & 96.63 & 98.19 \\
 \hline
\end{tabular}
\end{center}
\end{table}

\begin{table}
\begin{center}
\caption{Performance of party 1's models using SFE, CTFE, LTFE, and Non-Cooperative (NC) case on synthetic data with 30\% of sharing. A, P, R, and F1 correspond to accuracy, precision, recall, and F1 score, respectively. The results are shown for labels 7, 8, and 9 for brevity and consistency.}
\label{table:synth_performance} 
\begin{tabular}{c|cccccc}
\hline
Method & Label & A  & P & R & F1 \\
\hline
 \multirow{3}{*}{NC}
& SEVEN & 94.05 & 73.84 & 63.60 & 68.33\\
& EIGHT & 93.75 & 69.80 & 64.90 & 67.26\\
& NINE  & 94.06 & 71.58 & 65.06 & 68.17\\
 \hline
  \multirow{3}{*}{SFE}
& SEVEN & 94.10 & 77.62 & 58.40 & 66.65\\
& EIGHT & 93.65 & 75.04 & 53.67 & 62.58\\
& NINE  & 93.98 & 77.15 & 54.58 & 63.93\\
 \hline
\multirow{3}{*}{CTFE}
& SEVEN & 93.96 & 74.88 & 60.53 & 66.94\\
& EIGHT & 94.45 & 76.34 & 63.48 & 69.32\\
& NINE  & 94.39 & 73.97 & 65.83 & 69.66\\
 \hline
\multirow{3}{*}{LTFE}
& SEVEN & 95.94 & 83.39 & 74.59 & 78.75\\
& EIGHT & 96.05 & 81.59 & 77.54 & 79.51\\
& NINE  & 95.70 & 80.50 & 73.91 & 77.07\\
 \hline
\end{tabular}
\end{center}
\end{table}

\begin{table}
\begin{center}
\caption{Performance of party 1's models using SFE, CTFE, LTFE, and Non-Cooperative (NC) case on fraud detection data with 30\% of sharing. A, P, R, and F1 correspond to accuracy, precision, recall, and F1 score, respectively. The results are shown for fraudulent labels.}
\label{table:fraud_performance} 
\begin{tabular}{c|cccccc}
\hline
Method & Label & A  & P & R & F1 \\
\hline
NC  & FRAUDULENT & 97.73 & 94.12 & 80.00 & 86.49\\
 \hline
SFE & FRAUDULENT & 97.73 & 89.47 & 85.00 & 87.18\\
 \hline
CTFE & FRAUDULENT & 98.00 & 87.5 & 91.00 & 89.22\\
 \hline
LTFE & FRAUDULENT & 98.18 & 91.67 & 88.00 & 89.8\\
 \hline
\end{tabular}
\end{center}
\end{table}

\begin{table}
\begin{center}
\caption{Training time (in minutes) of SFE, LTFE, and CTFE on MNIST, synthetic data, and fraud detection dataset. The numbers correspond to one party and 10 epochs of training. Training time in secure and non-secure modes and the ratio between these two modes are reported.}
\label{table:time} 
\begin{tabular}{cc|cccc}
\hline
\multirow{2}{*}{Dataset} &  Model  & \multirow{2}{*}{Method}  &  Time &  Time  & Time \\
 & Type & & Secure & Non-Sec. & Ratio \\
\hline
\multirow{4}{*}{MNIST} & \multirow{4}{*}{FCN}  & SFE   & 54  & 2.2  & 24.5 \\
 &  & LTFE   & 337 & 2.8 & 120.3  \\
 &  & CTFE   & 207   & 0.8 & 258.7 \\
\hline
\multirow{4}{*}{MNIST} & \multirow{4}{*}{LeNet-5}  & SFE   & 814 & 2.4 & 339.1   \\
 &  & LTFE   & 2163  & 5.3 & 408.1 \\
 &  & CTFE   & 3344   & 3.1 & 1078.7 \\
\hline
 \multirow{4}{*}{Synthetic} & \multirow{4}{*}{FCN}   & SFE & 45 & 1.7 & 26.4 \\
 &  & LTFE  & 206  & 2.2 & 93.6 \\
 &  & CTFE  & 171  & 0.5 & 342   \\
 \hline
\multirow{4}{*}{Fraud} & \multirow{4}{*}{FCN}   & SFE & 20 & 0.33 & 60 \\
 &  & LTFE  & 80.9  & 0.06 & 1348.34\\
 &  & CTFE  & 70.7  & 0.08 & 883.75 \\
 \hline
\end{tabular}
\end{center}
\end{table}

\subsection{Information Leakage Using Membership Inference Attack}
\begin{figure*}
	\centering
	\subfloat[]{\includegraphics[width=0.25\linewidth]{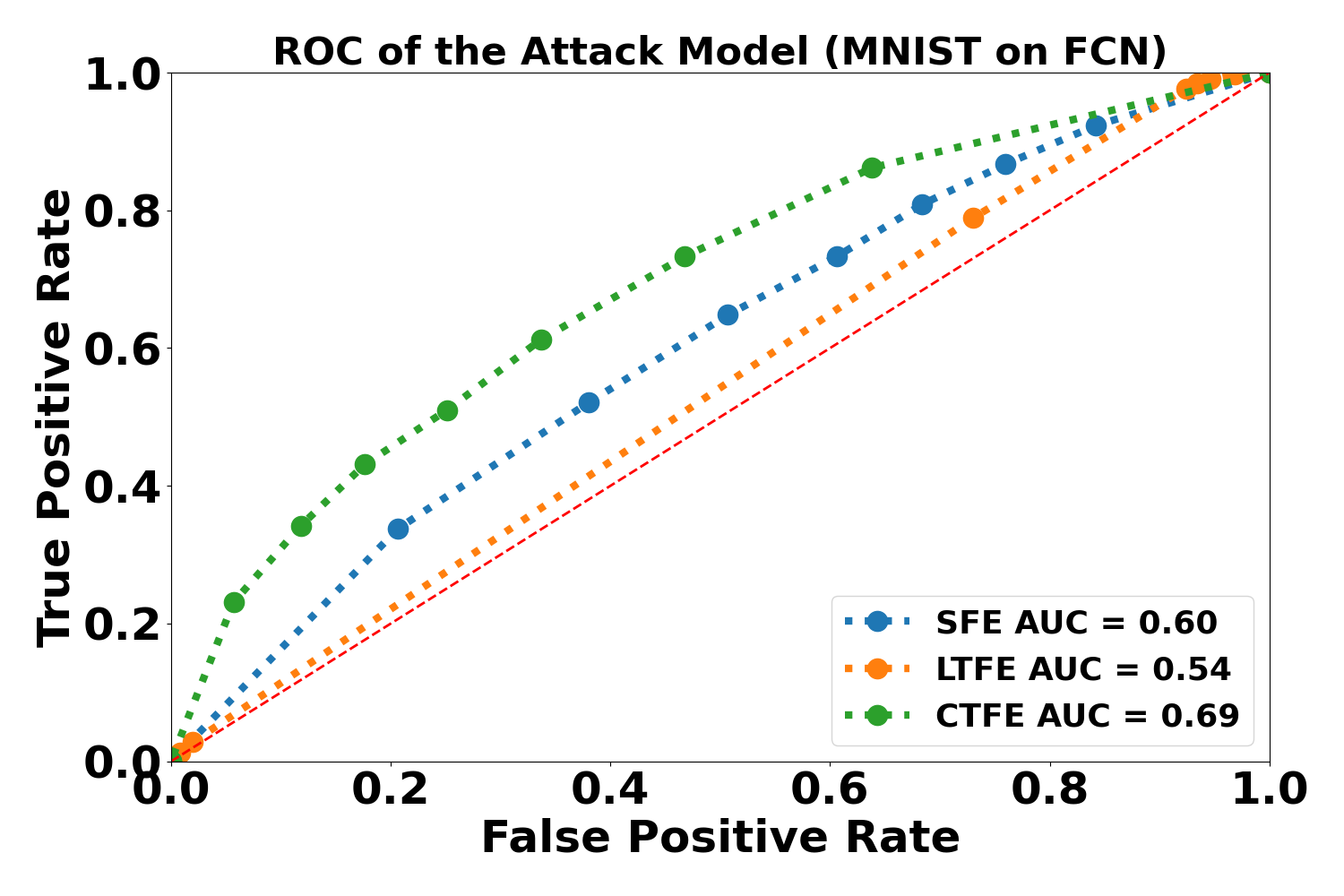}\label{a}}
	\hfill
	\subfloat[]{\includegraphics[width=0.25\linewidth]{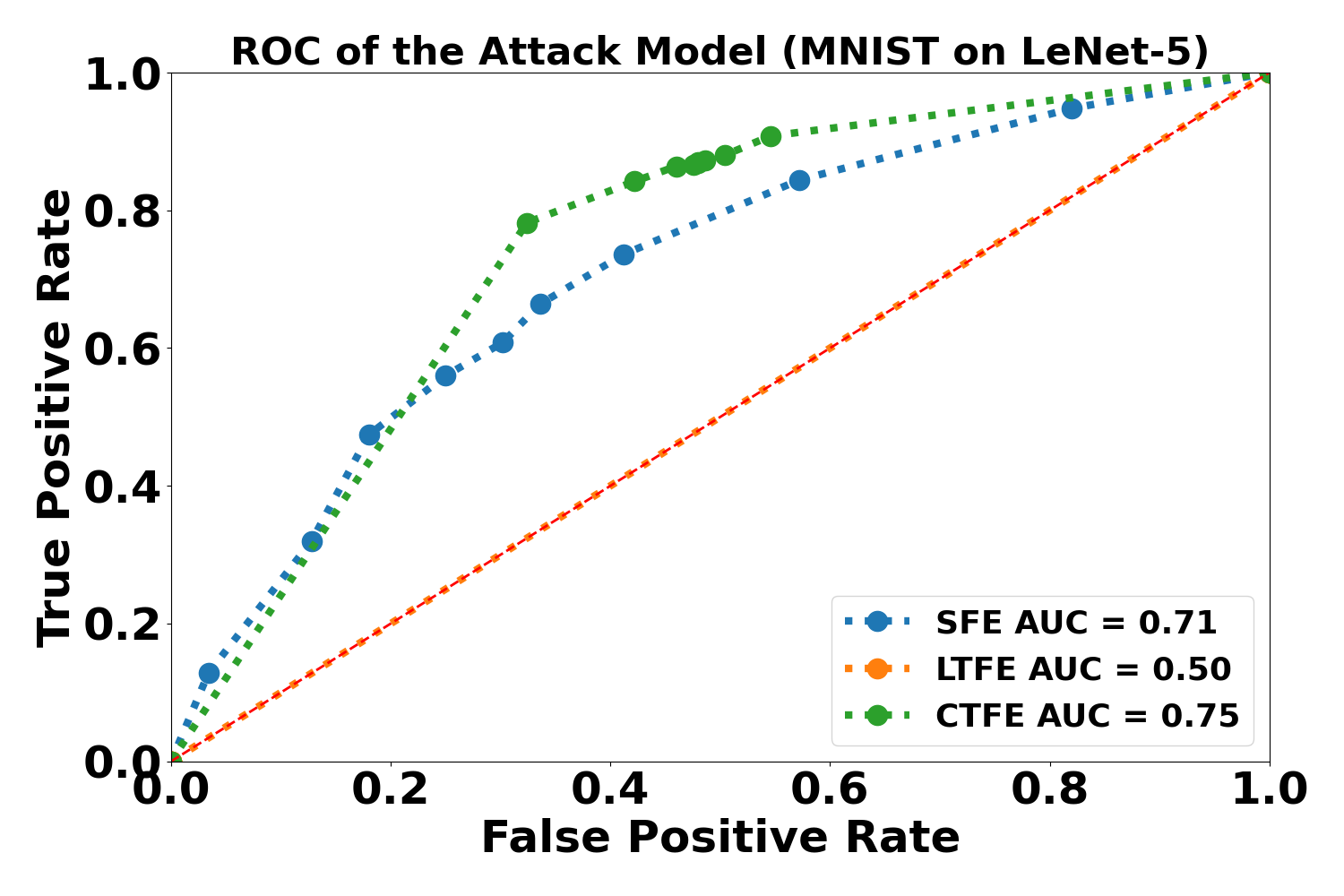}\label{b}}
	\hfill
	\subfloat[]{\includegraphics[width=0.25\linewidth]{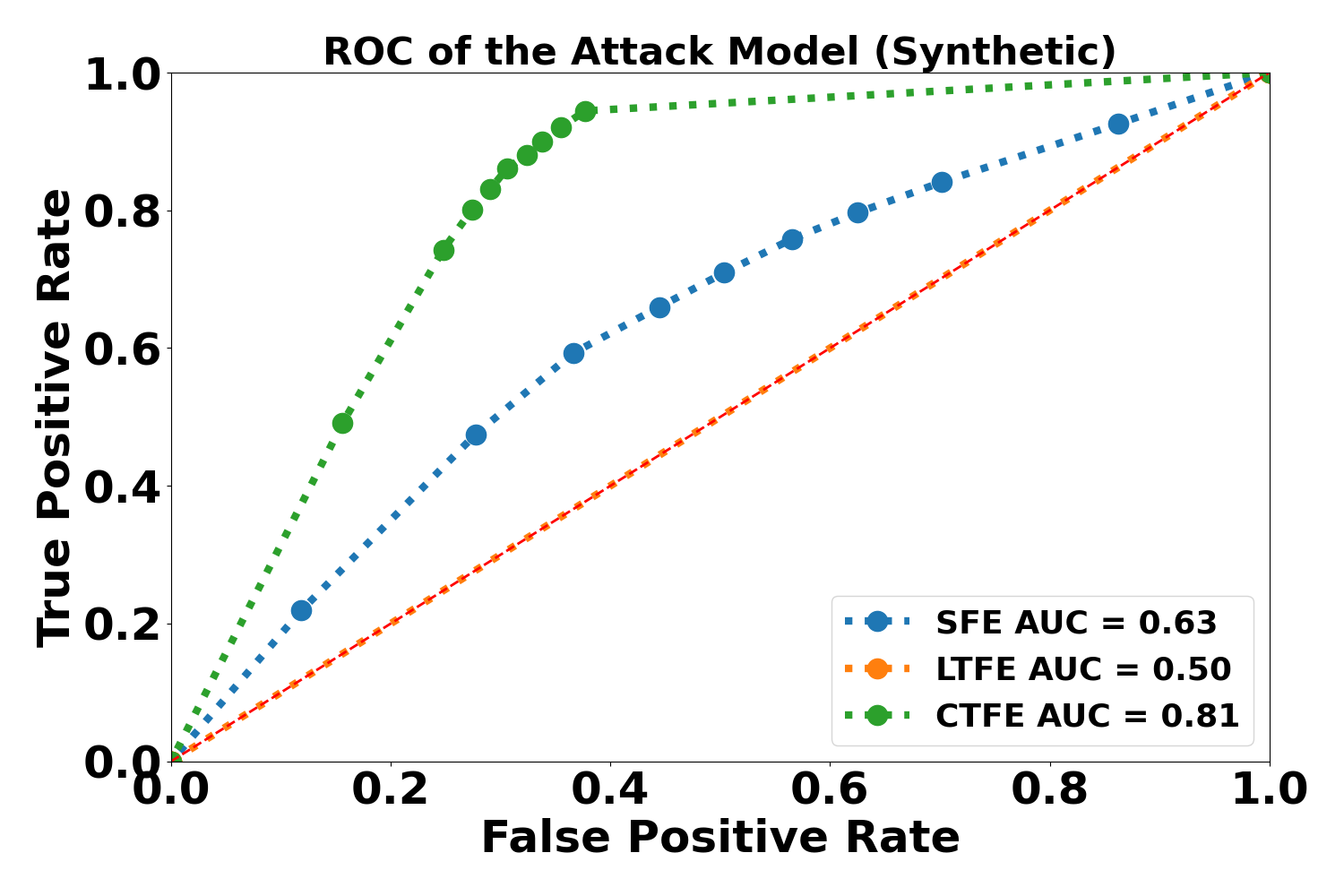}\label{b}}
	\hfill
	\subfloat[]{\includegraphics[width=0.25\linewidth]{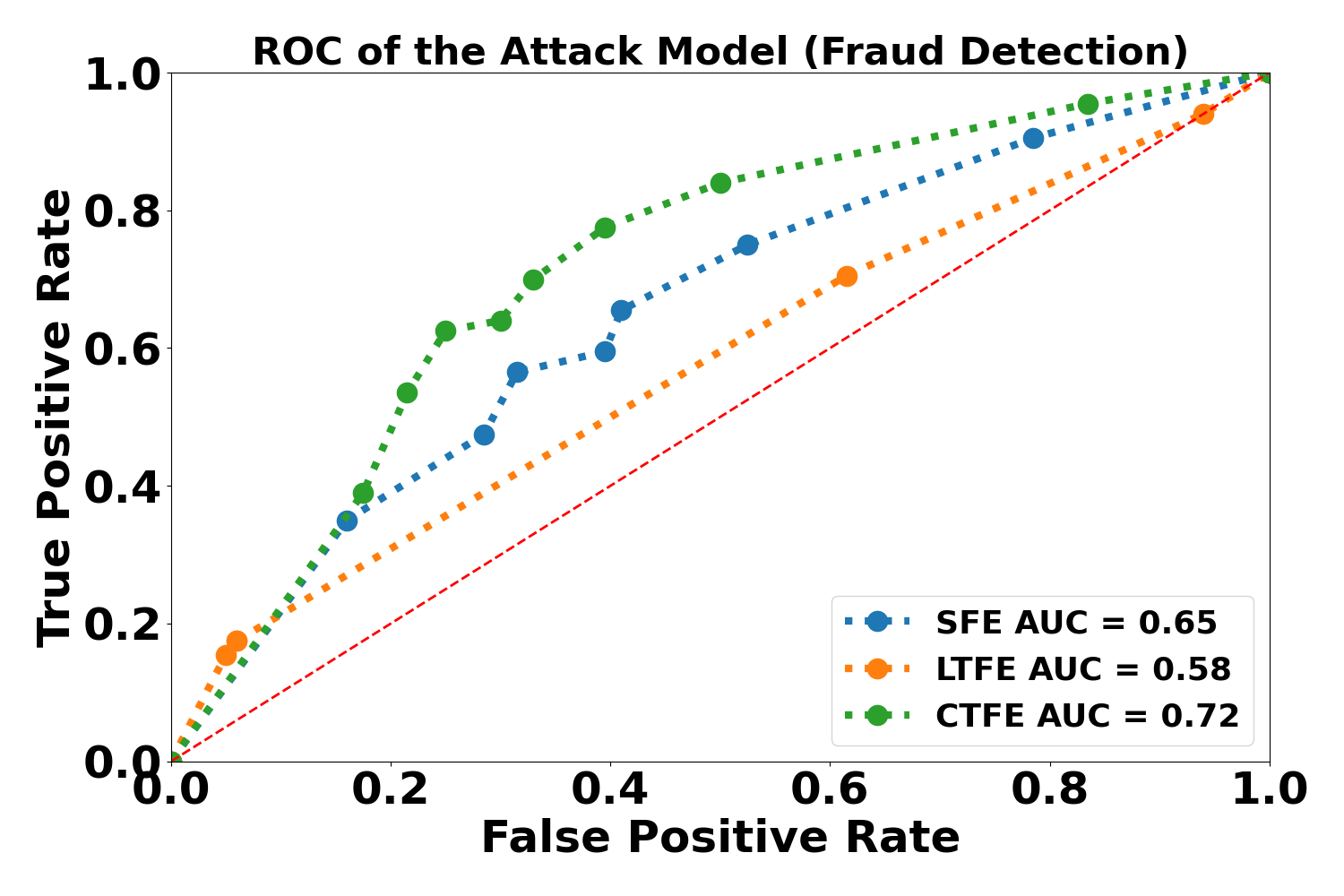}\label{b}}
	\caption{From left: The attack's ROC curve for the 3 methods on the MNIST dataset on FCN, MNIST dataset on LeNet-5, synthetic dataset, and fraud detection dataset.}
	\label{fig:roc} 
\end{figure*}

\begin{figure*}
	\centering
	\subfloat[]{\includegraphics[width=0.33\linewidth]{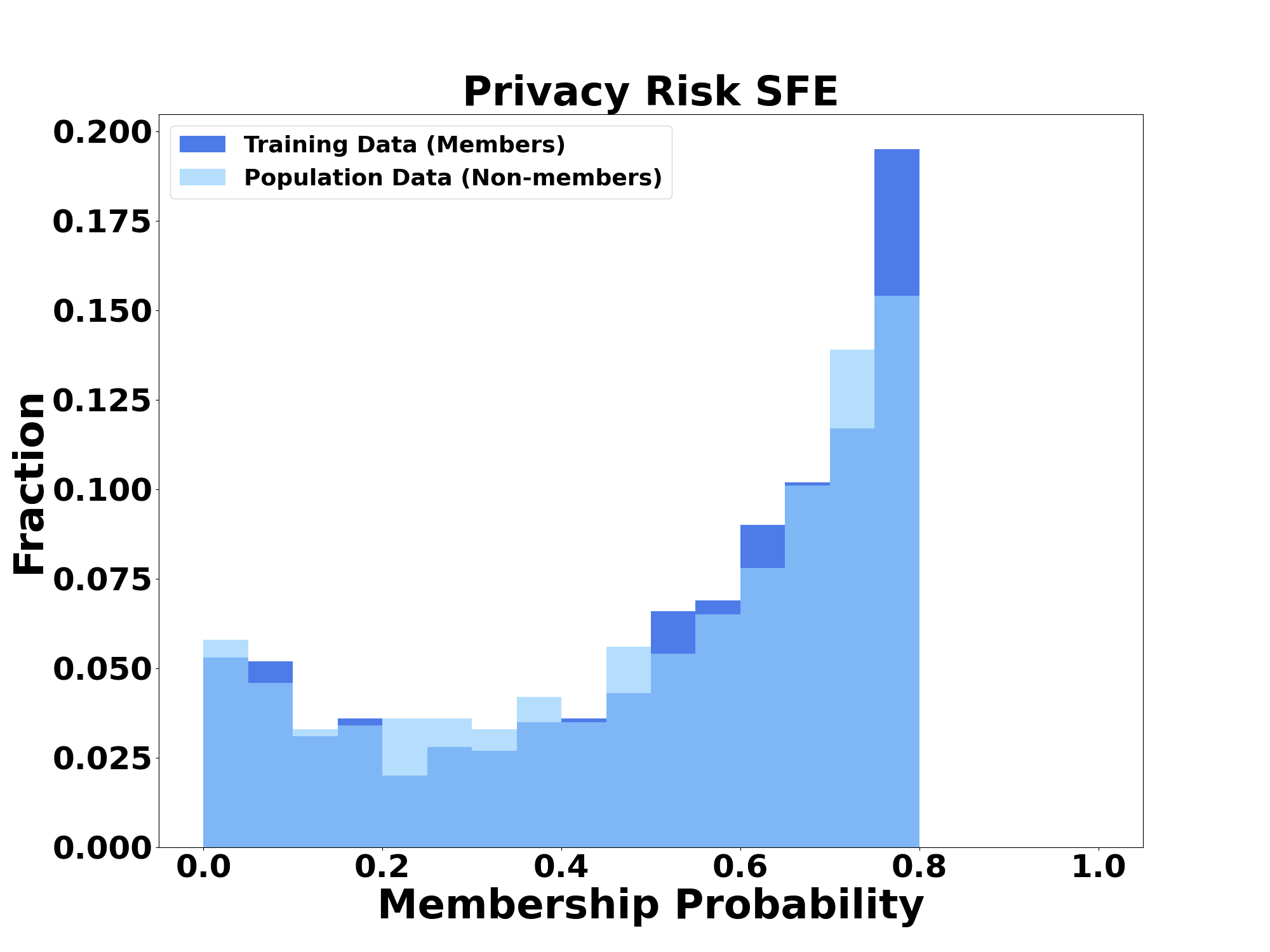}\label{a}}
	\hfill
	\subfloat[]{\includegraphics[width=0.33\linewidth]{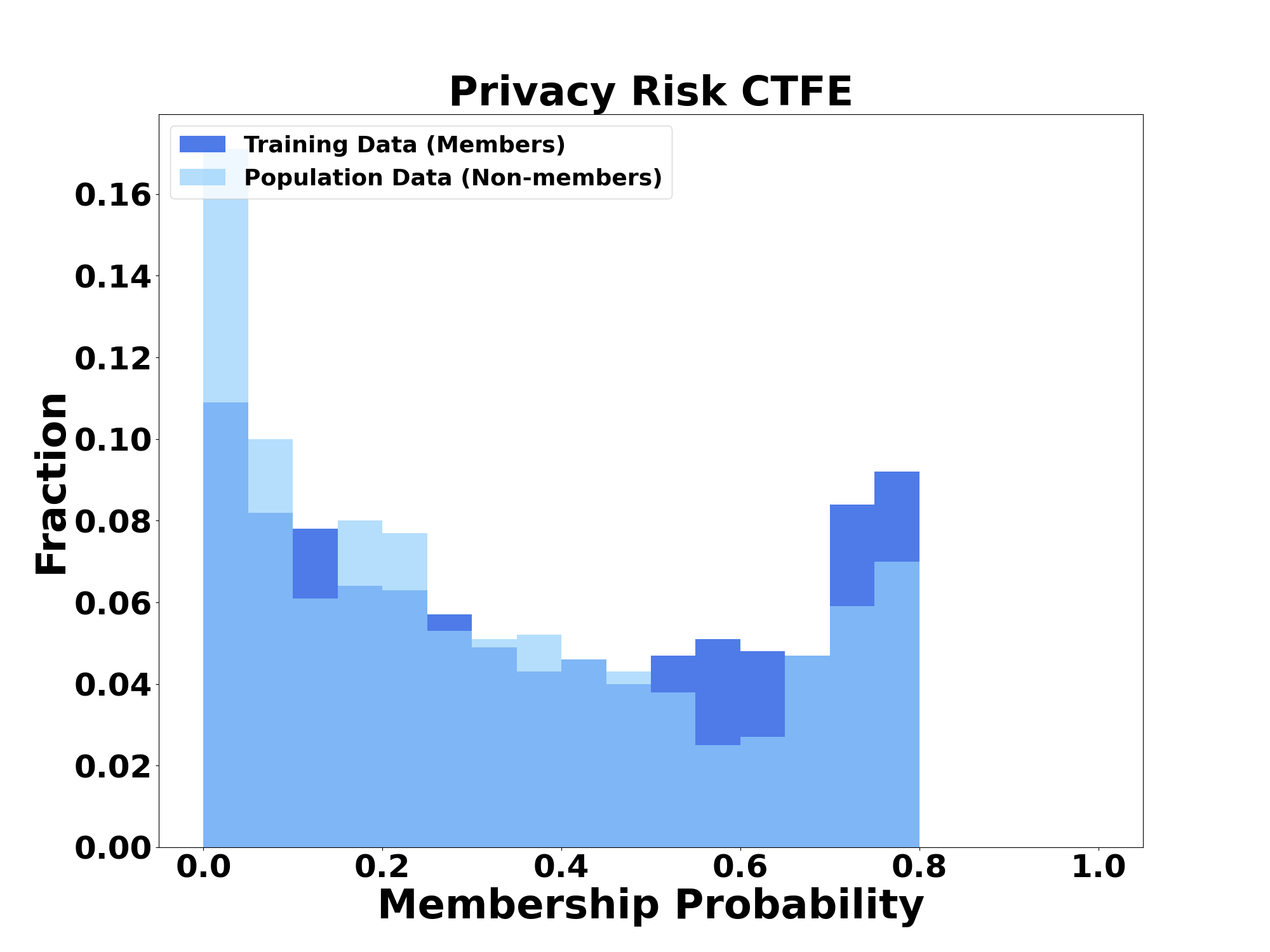}\label{b}}
	\hfill
	\subfloat[]{\includegraphics[width=0.33\linewidth]{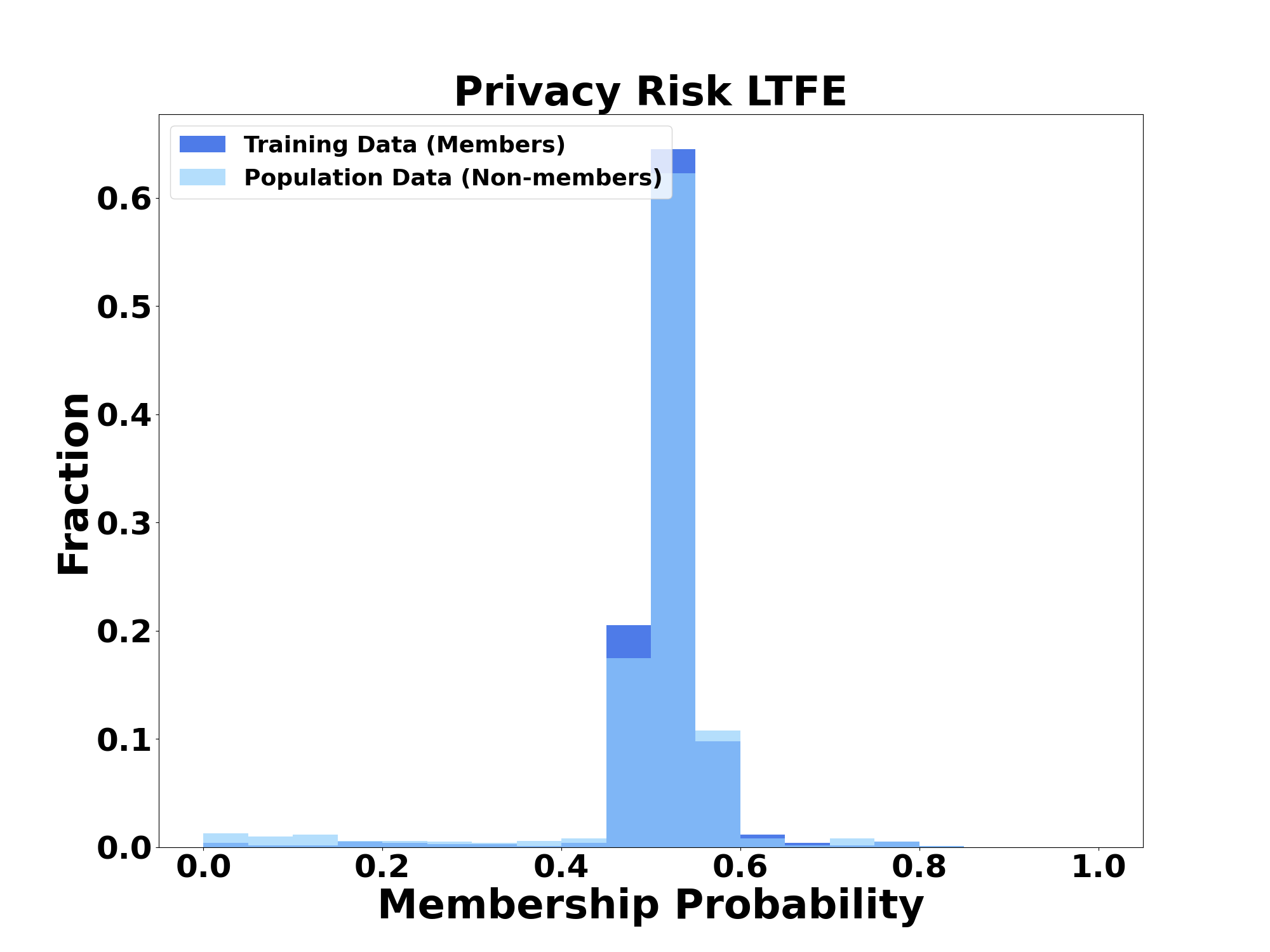}\label{c}}
	\\
	\subfloat[]{\includegraphics[width=0.33\linewidth]{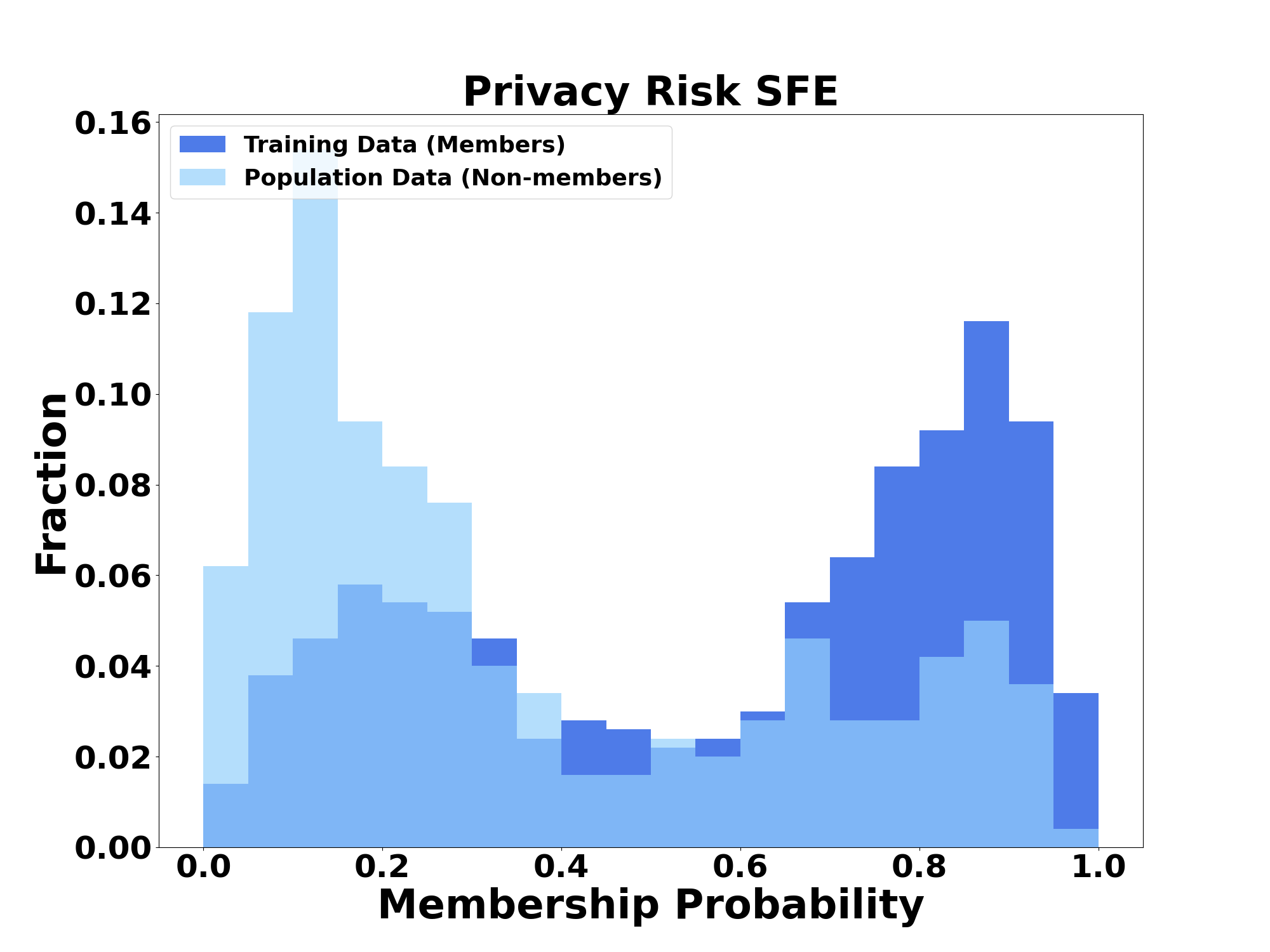}\label{a}}
	\hfill
	\subfloat[]{\includegraphics[width=0.33\linewidth]{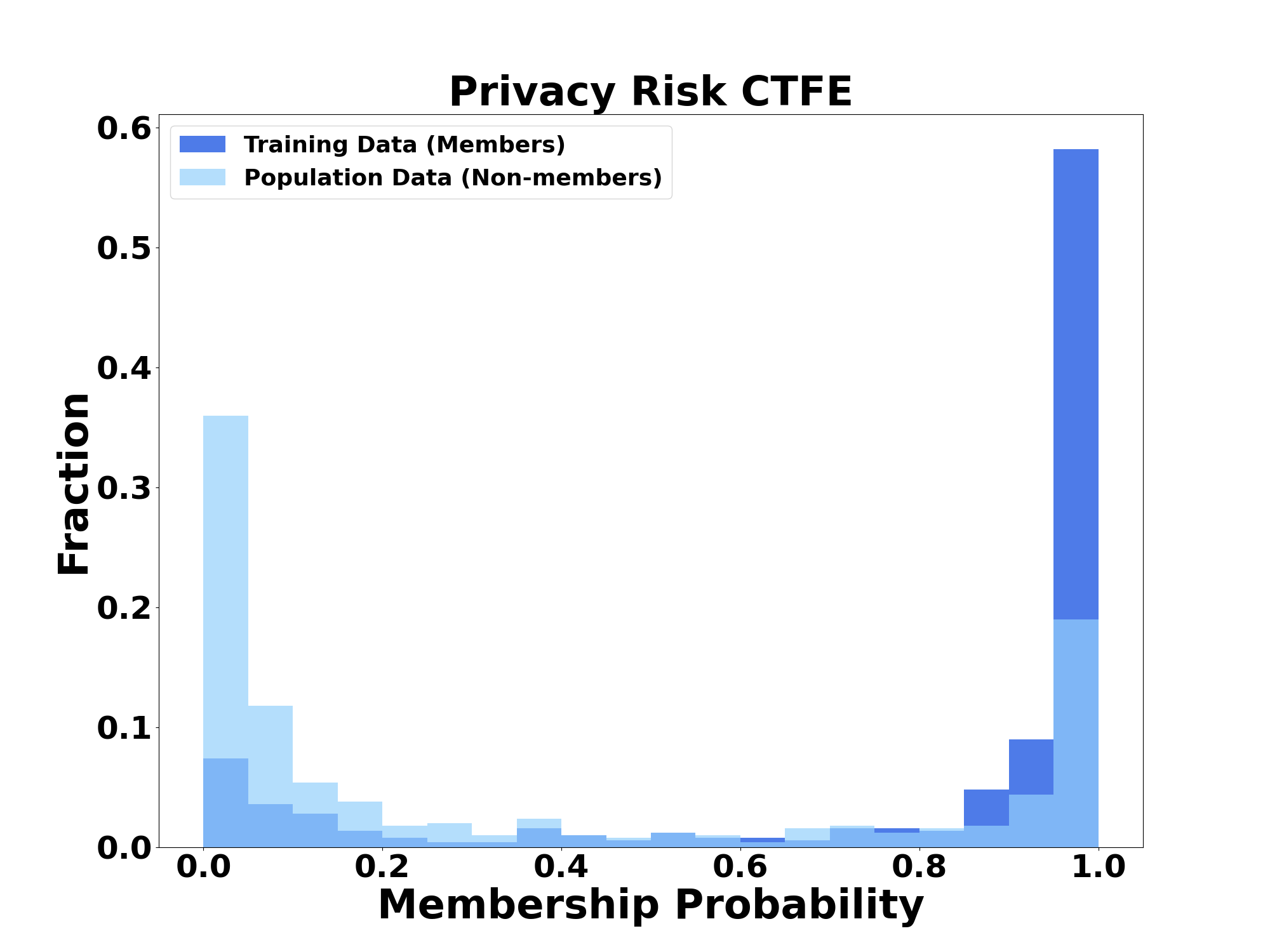}\label{b}}
	\hfill
	\subfloat[]{\includegraphics[width=0.33\linewidth]{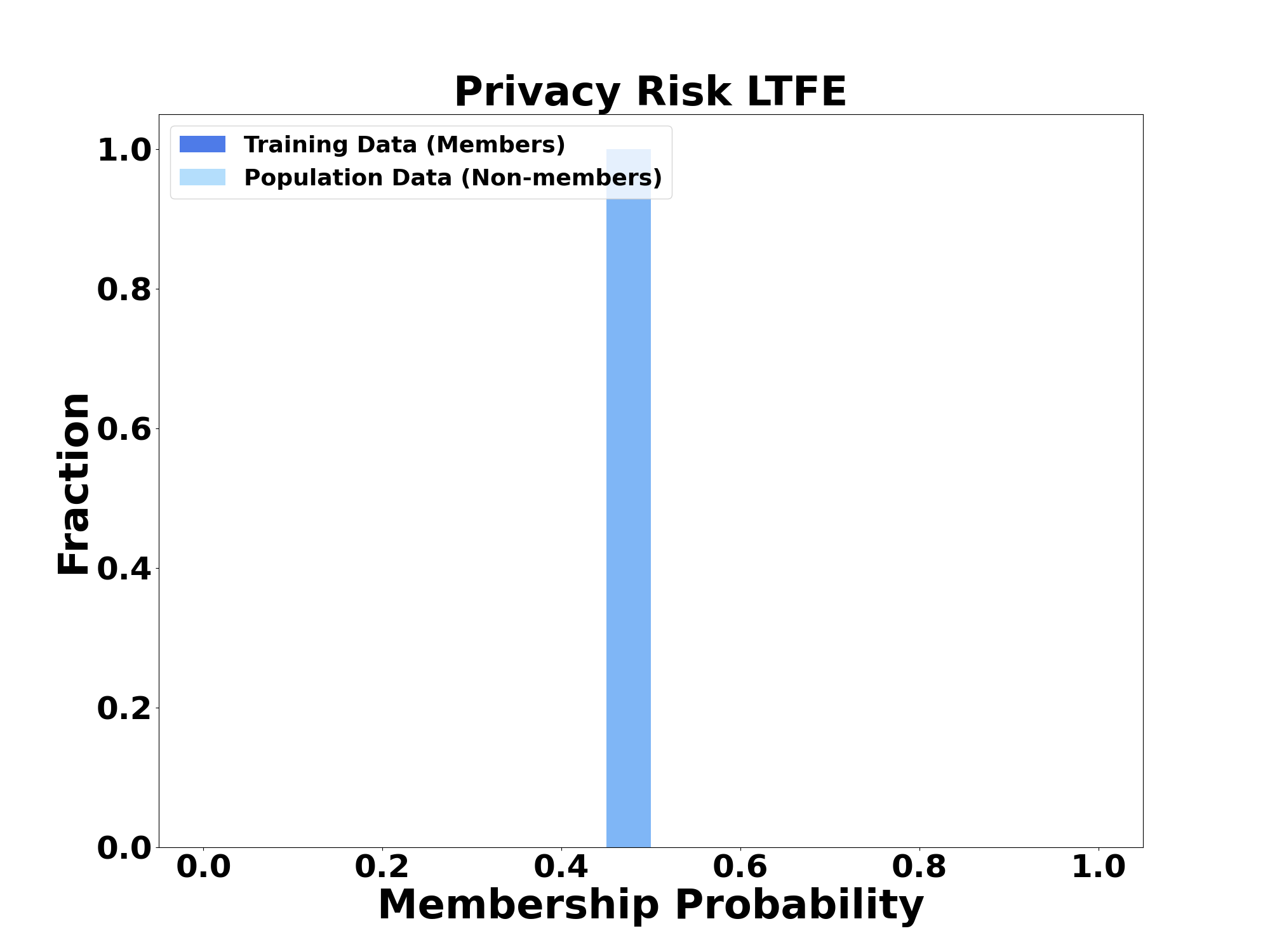}\label{c}}
	\\
	\subfloat[]{\includegraphics[width=0.33\linewidth]{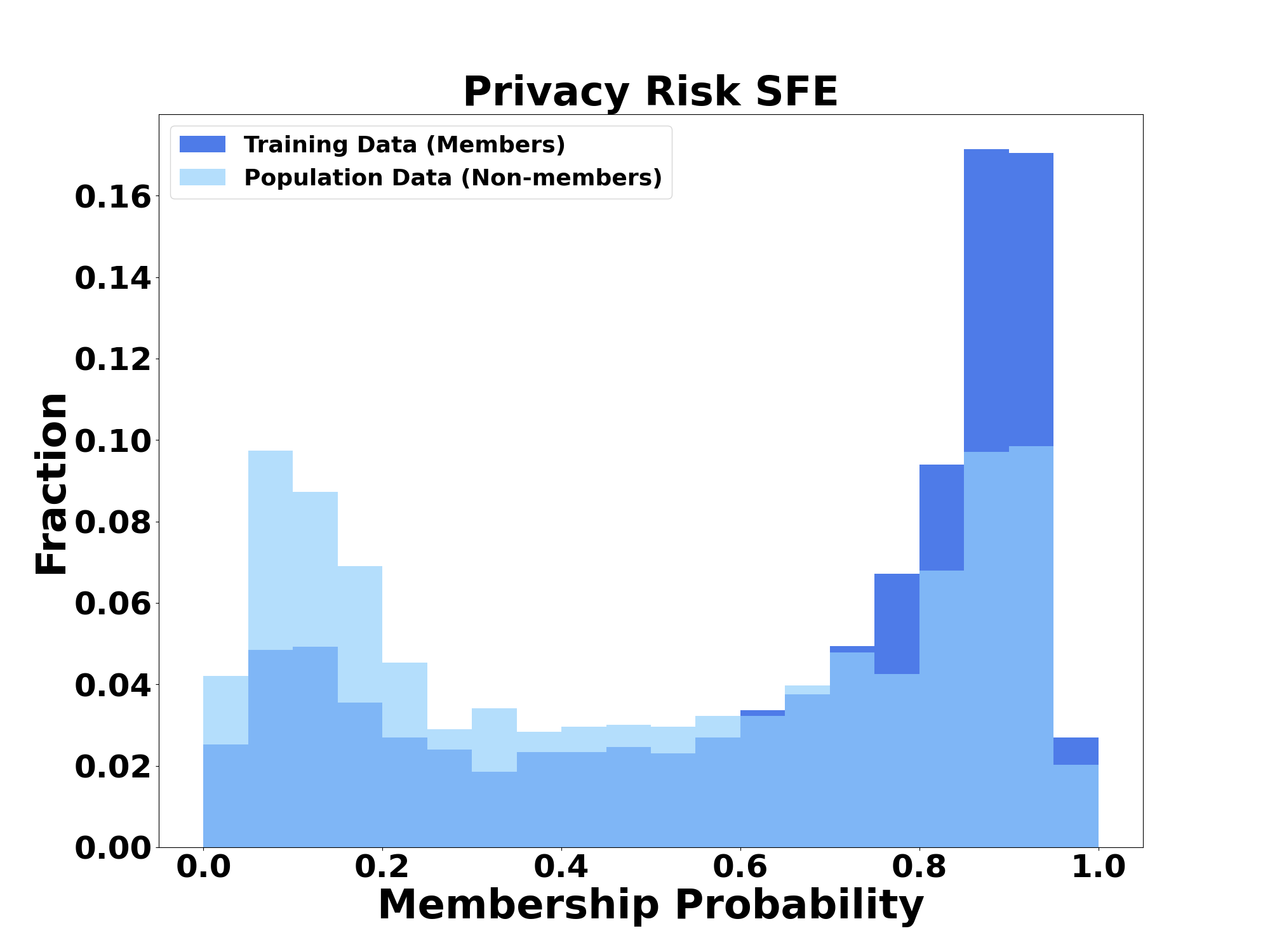}\label{a}}
	\hfill
	\subfloat[]{\includegraphics[width=0.33\linewidth]{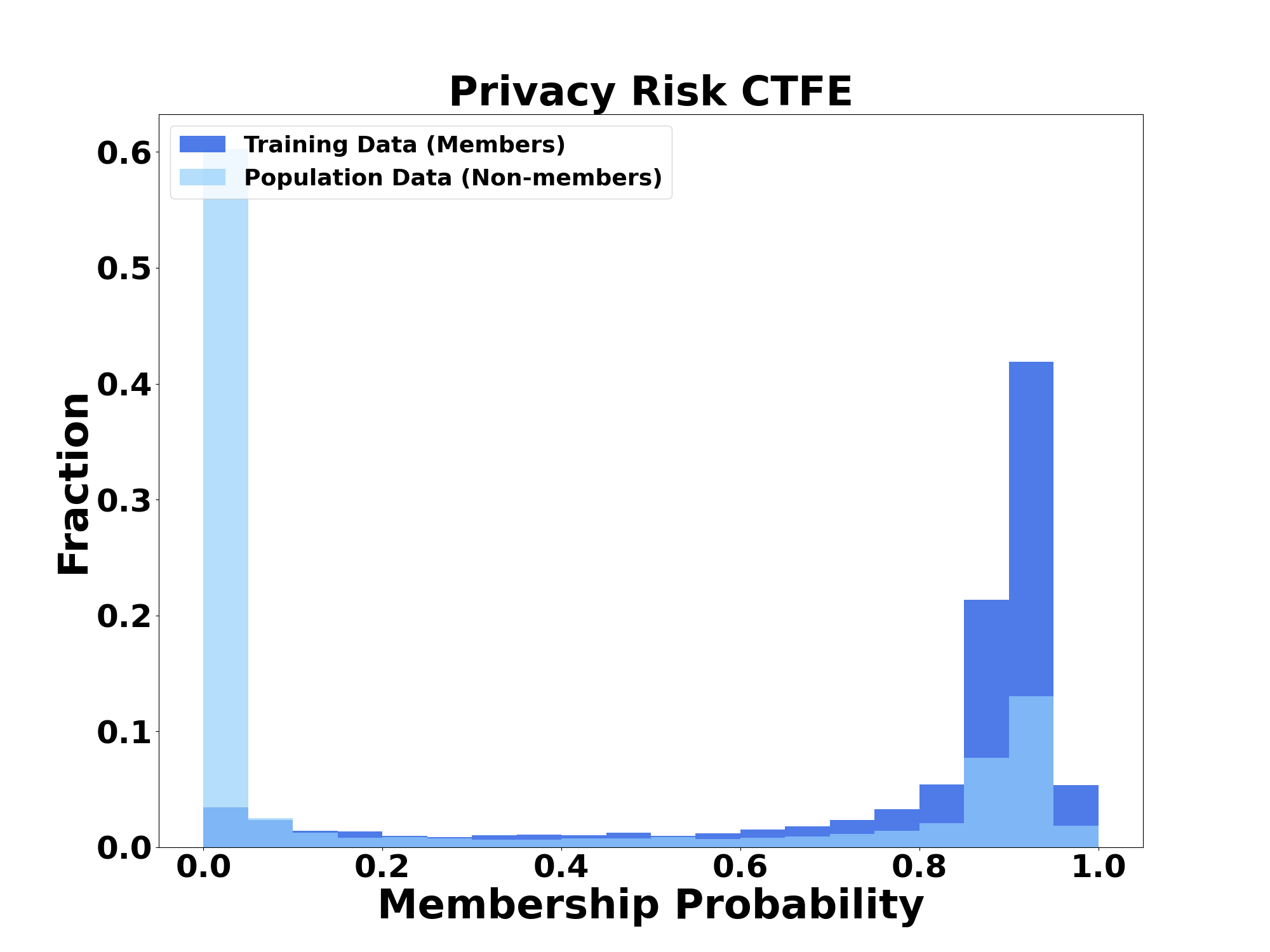}\label{b}}
	\hfill
	\subfloat[]{\includegraphics[width=0.33\linewidth]{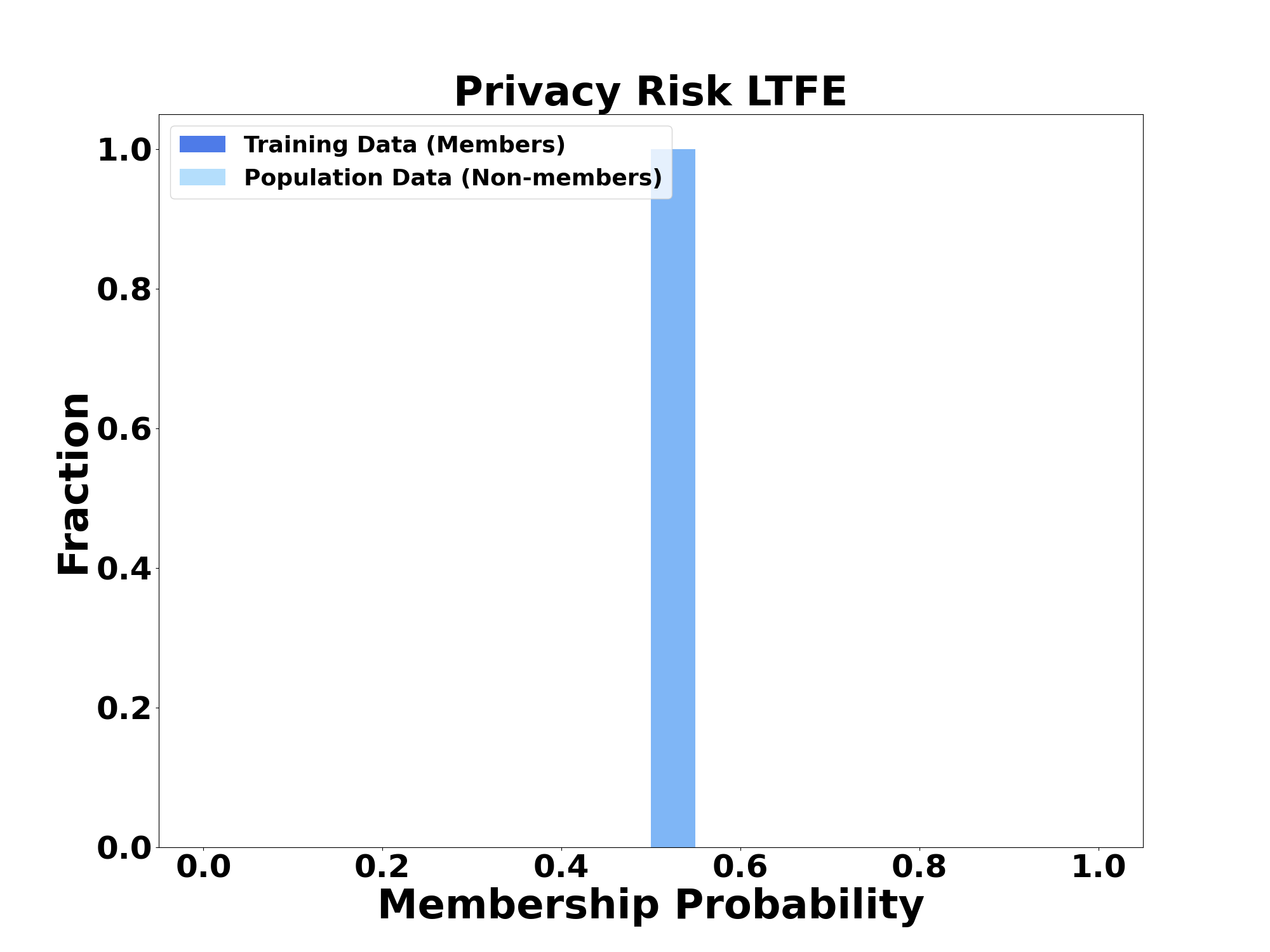}\label{c}}
	\\
	\subfloat[]{\includegraphics[width=0.33\linewidth]{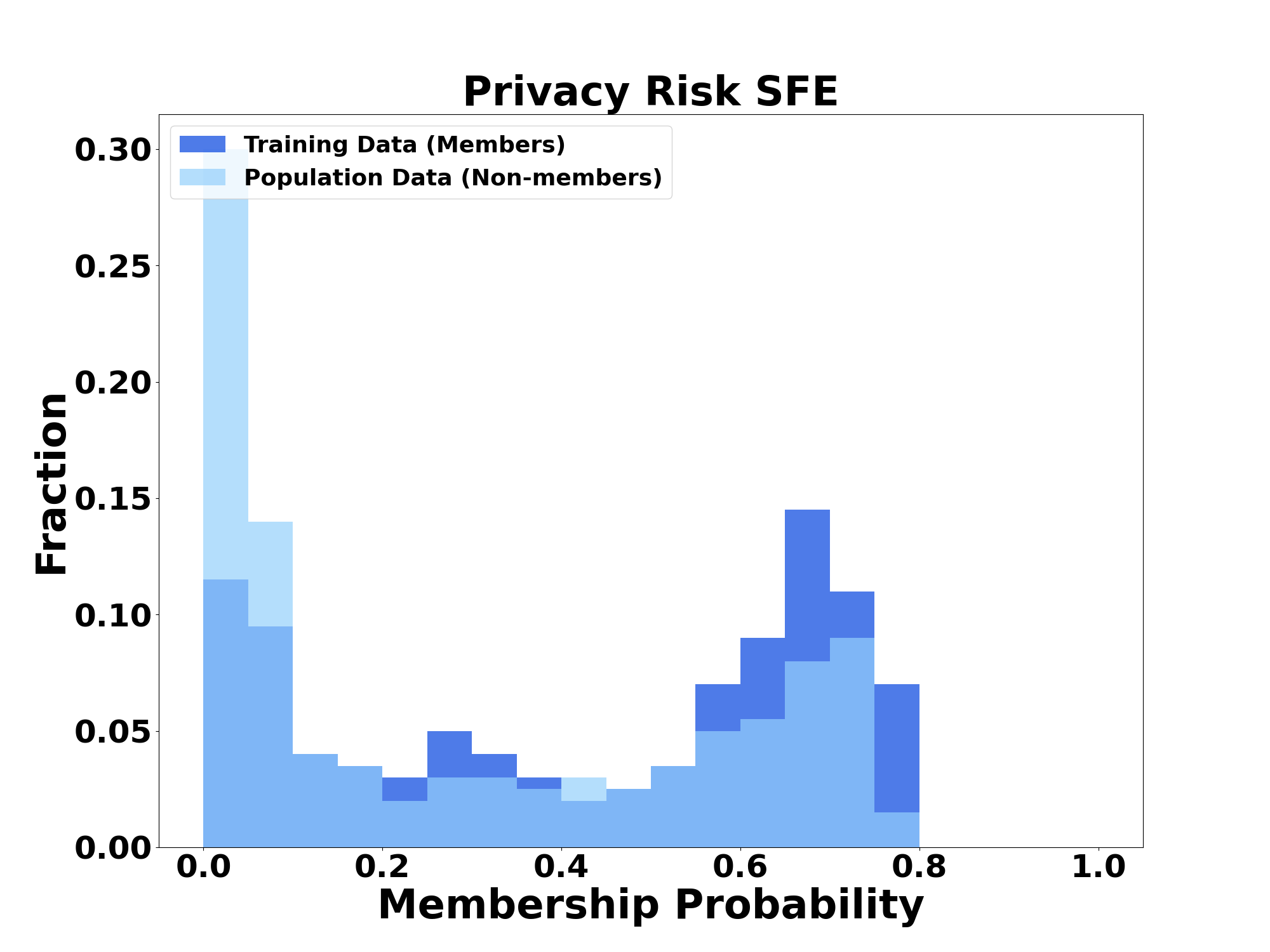}\label{a}}
	\hfill
	\subfloat[]{\includegraphics[width=0.33\linewidth]{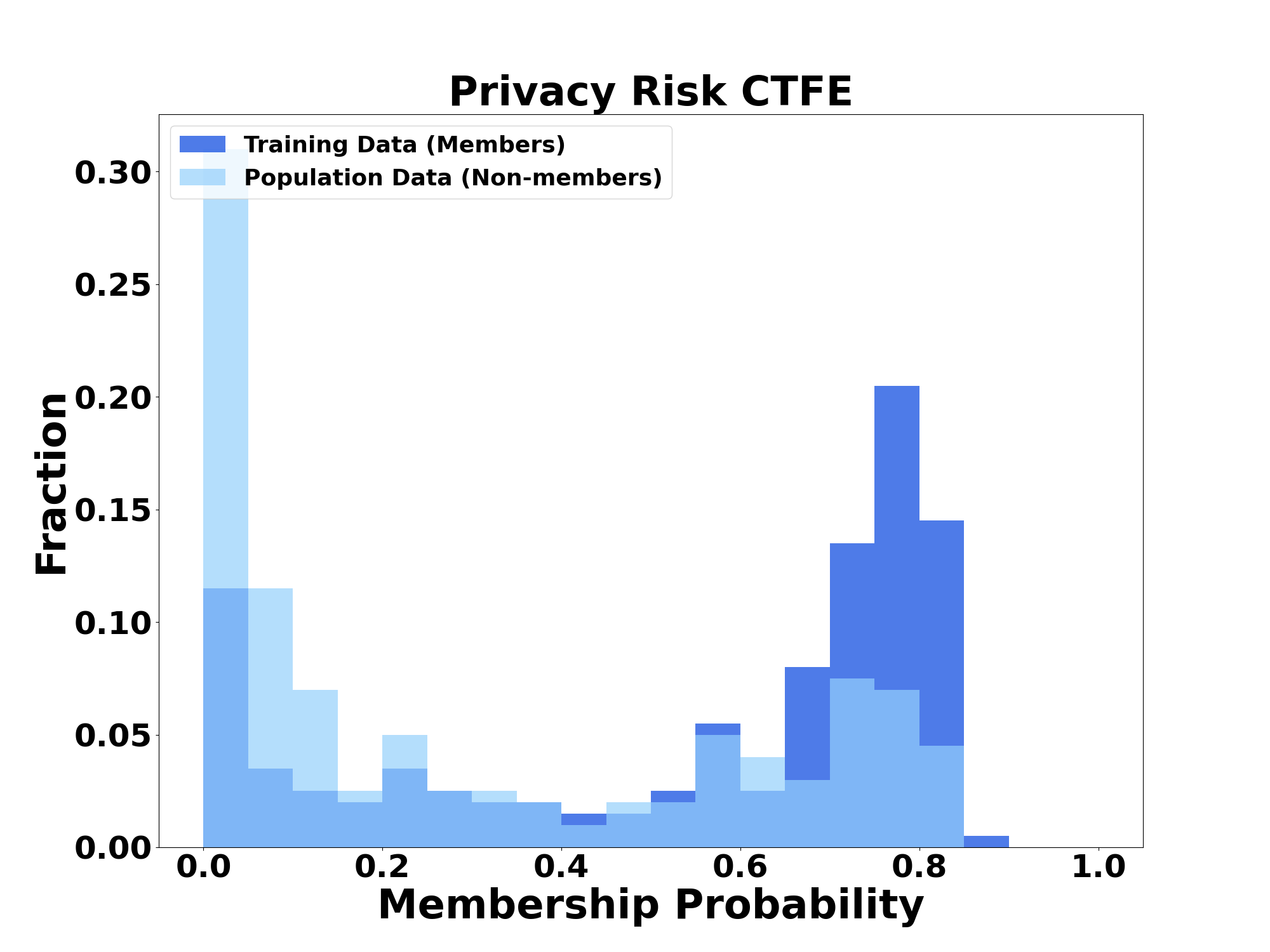}\label{b}}
	\hfill
	\subfloat[]{\includegraphics[width=0.33\linewidth]{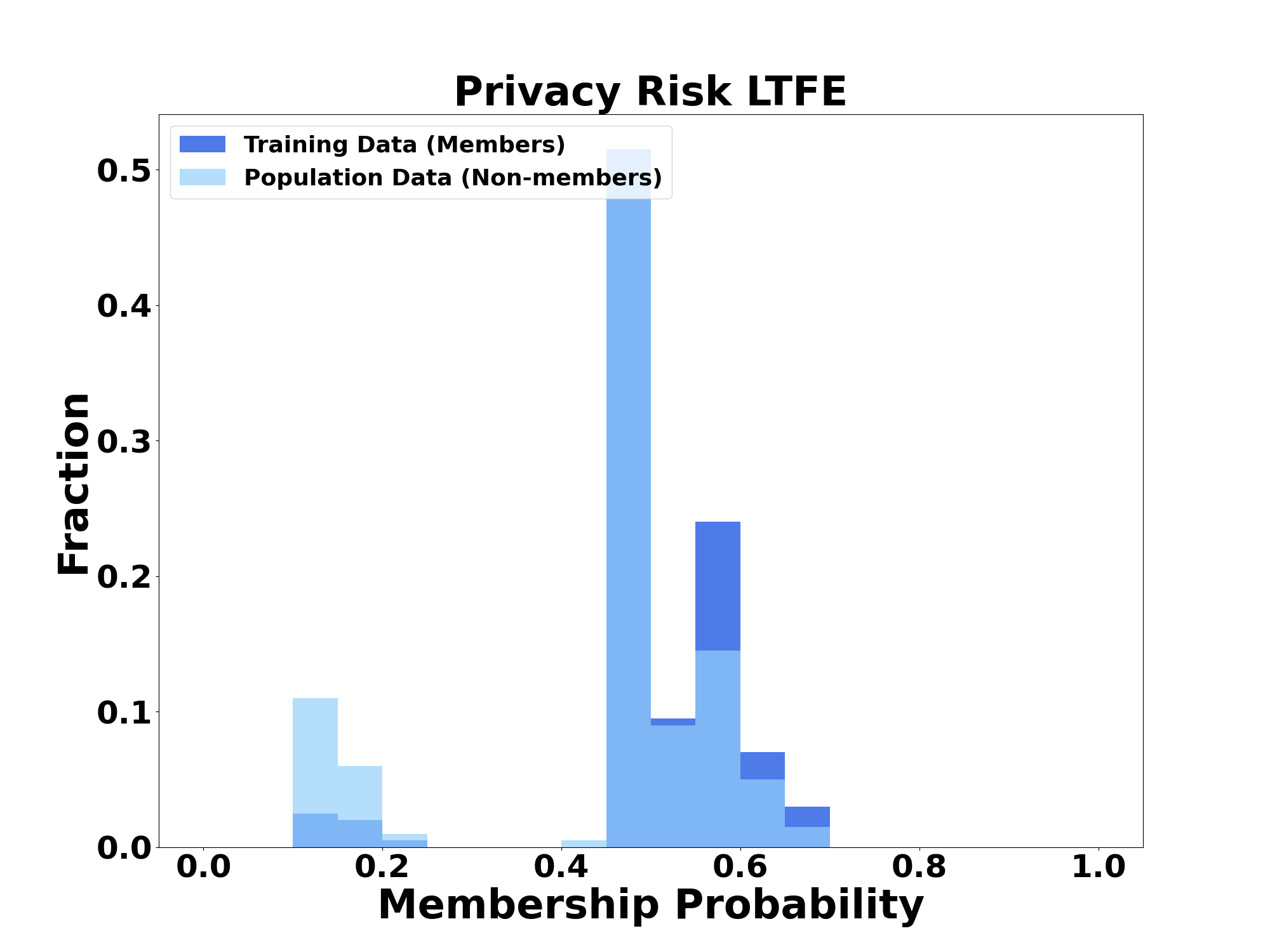}\label{c}}	
	\caption{Privacy risk of each method on MNIST dataset on FCN (top row), MNIST dataset on LeNet-5 (second row), synthetic dataset (third row), and fraud detection dataset (last row). From left: SFE, CTFE, and LTFE, respectively.}
	\label{fig:privacy_risk}
\end{figure*}

We evaluated the information leakage of the proposed scenarios in terms of the source of data by the method discussed in Sec. \ref{subsec:attack_models}. 
Prediction uncertainty is a measure of information leakage \cite{shokri2017membership}. A method with high uncertainty in its prediction reflects the fact that the model leaks less information about the samples. In Fig. \ref{fig:privacy_risk}, membership probabilities for the MNIST dataset are shown for member and non-member samples for the scenarios where label 1 corresponds to the member samples. The member samples probabilities are shown with dark blue and non-member samples with pale blue. In (c), (f), and (i) in Fig. \ref{fig:privacy_risk}, 50\% membership probability implies that the model leaks no information since  the attacker does not obtain any idea of whether a sample was in the training data or not (i.e., effectively a coin toss). If a model leaks a lot of information, the attack model will output a high membership probability for most training samples and a low membership probability for most non-training samples (i.e., (b), (e), and (h) in Fig. \ref{fig:privacy_risk}). It could be seen that the attack models against LTFE are highly uncertain about the samples' membership, whereas the attack models against  CTFE assign a high probability to training samples and a low probability to non-training samples. The attack models against SFE show a level of uncertainty less than the models against LTFE but greater than the models against CTFE.  Additionally, the Receiver Operating Characteristic (ROC) curves of the attack models are plotted in Fig. \ref{fig:roc}. A larger Area Under the ROC curve (AUC) means that the model leaks more information. We can see that the attack models against LTFE have the smallest AUC, meaning that the attack models have the highest uncertainty and LTFE leaks the least information. The attack models against CTFE have the largest AUC, meaning that the attack models have the lowest uncertainty and CTFE leaks the most information. The amount of information leaked by SFE is between those of LTFE and CTFE. Figs. \ref{fig:roc} and \ref{fig:privacy_risk} are consistent with each other.

\subsection{Timing}
The training time for the proposed methods is reported in Table~\ref{table:time} for MNIST FCN, MNIST LeNet-5, and synthetic dataset cases. In this table, all the numbers are in minutes. Secure training time, non-secure training time, and the ratio of the secure training time to the non-secure training time are reported in Table~\ref{table:time}. It could be seen that for fully connected network architectures, LTFE is the slowest, and SFE is the fastest. However, for LeNet-5 which has a CNN architecture for the feature extractor, the CTFE is the slowest. This observation is consistent with our expectation of the computational cost of the CNN layers with respect to fully connected layers since CNN layers require more mathematical operations. We also empirically observed that the models trained using SFE, CTFE, and LTFE in non-secure mode have comparable accuracy to the models trained in the secure mode. Additionally, since training in non-secure mode sacrifices input data privacy,  the training speed in non-secure mode is faster than the speed in the secure mode.

\subsection{Performance, Privacy, and Speed Trade-Offs}
Based on our experimental results (i.e., from Tables~\ref{table:mnist_performance_fcn}, \ref{table:mnist_performance_lenet}, \ref{table:synth_performance}, \ref{table:fraud_performance} and \ref{table:time}; Figs.~\ref{fig:roc} and \ref{fig:privacy_risk}), the comparison of CTFE, SFE, and LTFE with respect to performance, privacy, and training speed is consistent with what we summarized in Fig.~\ref{fig:trade_off}. Table~\ref{table:tradeoffs} shows the comparison according to our experimental results. For accuracy, LTFE and CTFE have comparable results in most cases. However, we empirically observed that CTFE and SFE are more sensitive to training noise than LTFE. For example, the model weights may be stuck to a local optimum and cannot be further changed during the training process. Therefore, we conclude that LTFE has the overall best performance among the three methods. SFE has the least accuracy. The performance of CTFE is between LTFE's and SFE's. As for privacy, Fig.~\ref{fig:roc} and Fig.~\ref{fig:privacy_risk} show that LTFE leaks the least information, whereas CTFE leaks the most information. SFE has a medium level of privacy. As for the training speed, SFE is the fastest algorithm, whereas LTFE is typically the slowest according to experiments. CTFE is slower than SFE, however, it could be faster than LTFE based on the network architecture. In real-world cases, since each party will have its own machine, the training process can be parallel and therefore may take less time. To apply our algorithms for $n$-party (with $n>2$) scenarios, one possible solution is that each party collaborates with one other party at a time (i.e., pairwise collaborative learning) to train its model. Thus, the n-party case is converted to a general 2-party scenario. 

\begin{table}
\begin{center}
\caption{Comparisons among SFE, CTFE, and LTFE (relative speed of CTFE and LTFE depends on the network architecture).}
\label{table:tradeoffs} 
\begin{tabular}{cccc}
\hline\noalign{\smallskip}
Algorithm & Accuracy & Privacy  & Speed  \\
\hline
SFE & Lowest & Medium & Fastest \\
CTFE & Medium & Worst & Slower than SFE \\
LTFE & Highest & Safest & Slower than SFE \\
 \hline
\end{tabular}
\end{center}
\end{table}

\subsection{Comparative Study}
In this section, we highlight the differences between our method and commonly used methods in the literature. In FL \cite{MMRHYB17}, there is a model, and each local party updates the weights of the model using its own data, and the nodes send the weights and their gradients to the server. However, in our work, collaborative learning is based on the SMPC approach in which the privacy of the inputs is protected. Also, each party has its own model. In \cite{ding2022privacy}, the proposed method sends the extracted features to the cloud instead of the raw data. These extracted features are not encrypted. Moreover, the focus of their work is model inference. However, our method addresses the privacy of the data during training. In \cite{cai2022privacy}, the authors proposed privacy-preserving CNN feature extraction on medical images. In their method, the data owner encrypts its data and sends it to the servers. The main difference between our method and their method is that they have assumed that they have the trained parameters of the CNN, while we are training the feature extractor in the secure domain. Also, in their method, they only have one data owner while we are addressing the case where we have two organizations cooperating in the training phase.

\section{Conclusion}
\label{sec.conclusion}
Two variants (SFE and LTFE) of feature-based privacy-preserving collaborative learning are proposed and compared with a baseline (CTFE).  The algorithms are evaluated on three datasets in terms of performance, privacy, and computational cost. The trade-offs between these methods have been discussed. LTFE is the most secure and accurate algorithm but is more computationally expensive. SFE is the least computationally expensive. CTFE is the baseline algorithm that has the least privacy-preserving capability. Prospective works can be focused on further improving the performance, accelerating the training speed, increasing the privacy of these three methods, and expanding to scenarios with more than 2 parties.

\bibliographystyle{IEEEtran}
\bibliography{references}

\begin{thebibliography}{10}
\providecommand{\url}[1]{#1}
\csname url@samestyle\endcsname
\providecommand{\newblock}{\relax}
\providecommand{\bibinfo}[2]{#2}
\providecommand{\BIBentrySTDinterwordspacing}{\spaceskip=0pt\relax}
\providecommand{\BIBentryALTinterwordstretchfactor}{4}
\providecommand{\BIBentryALTinterwordspacing}{\spaceskip=\fontdimen2\font plus
\BIBentryALTinterwordstretchfactor\fontdimen3\font minus
  \fontdimen4\font\relax}
\providecommand{\BIBforeignlanguage}[2]{{%
\expandafter\ifx\csname l@#1\endcsname\relax
\typeout{** WARNING: IEEEtran.bst: No hyphenation pattern has been}%
\typeout{** loaded for the language `#1'. Using the pattern for}%
\typeout{** the default language instead.}%
\else
\language=\csname l@#1\endcsname
\fi
#2}}
\providecommand{\BIBdecl}{\relax}
\BIBdecl

\bibitem{MMRHYB17}
B.~McMahan, E.~Moore, D.~Ramage, S.~Hampson, and B.~A. y~Arcas,
  ``Communication-efficient learning of deep networks from decentralized
  data,'' in \emph{Proceedings of the International Conference on Artificial
  Intelligence and Statistics}, Lauderdale, FL, Apr. 2017, pp. 1273--1282.

\bibitem{KJMYRSB16}
J.~Kone{\v{c}}n{\`y}, H.~B. McMahan, F.~X. Yu, P.~Richt{\'a}rik, A.~T. Suresh,
  and D.~Bacon, ``Federated learning: Strategies for improving communication
  efficiency,'' \emph{arXiv preprint arXiv:1610.05492}, 2016.

\bibitem{NSH19}
M.~Nasr, R.~Shokri, and A.~Houmansadr, ``Comprehensive privacy analysis of deep
  learning: Passive and active white-box inference attacks against centralized
  and federated learning,'' in \emph{Proceedings of the IEEE Symposium on
  Security and Privacy}, Francisco, CA, May 2019, pp. 739--753.

\bibitem{song2020analyzing}
M.~Song, Z.~Wang, Z.~Zhang, Y.~Song, Q.~Wang, J.~Ren, and H.~Qi, ``Analyzing
  user-level privacy attack against federated learning,'' \emph{IEEE Journal on
  Selected Areas in Communications}, vol.~38, no.~10, pp. 2430--2444, 2020.

\bibitem{wang2019beyond}
Z.~Wang, M.~Song, Z.~Zhang, Y.~Song, Q.~Wang, and H.~Qi, ``Beyond inferring
  class representatives: User-level privacy leakage from federated learning,''
  in \emph{Proceedings of the IEEE Conference on Computer Communications},
  Paris, France, Apr. 2019, pp. 2512--2520.

\bibitem{damgaard2012multiparty}
I.~Damg{\aa}rd, V.~Pastro, N.~Smart, and S.~Zakarias, ``Multiparty computation
  from somewhat homomorphic encryption,'' in \emph{Proceedings of the Annual
  Cryptology Conference}, Santa Barbara, CA, Aug. 2012, pp. 643--662.

\bibitem{gordon2015constant}
S.~D. Gordon, F.-H. Liu, and E.~Shi, ``Constant-round mpc with fairness and
  guarantee of output delivery,'' in \emph{Proceedings of the Annual Cryptology
  Conference}, Santa Barbara, CA, Aug. 2015, pp. 63--82.

\bibitem{rabin1989verifiable}
T.~Rabin and M.~Ben-Or, ``Verifiable secret sharing and multiparty protocols
  with honest majority,'' in \emph{Proceedings of the Annual ACM Symposium on
  Theory of Computing}, Santa Barbara, CA, Aug. 1989, pp. 73--85.

\bibitem{pass2017formal}
R.~Pass, E.~Shi, and F.~Tramer, ``Formal abstractions for attested execution
  secure processors,'' in \emph{Proceedings of the Annual International
  Conference on the Theory and Applications of Cryptographic Techniques},
  Paris, France, Apr. 2017, pp. 260--289.

\bibitem{chillotti2017faster}
I.~Chillotti, N.~Gama, M.~Georgieva, and M.~Izabach{\`e}ne, ``Faster packed
  homomorphic operations and efficient circuit bootstrapping for tfhe,'' in
  \emph{Proceedings of the International Conference on the Theory and
  Application of Cryptology and Information Security}, Hong Kong, China, Dec.
  2017, pp. 377--408.

\bibitem{mukherjee2016two}
P.~Mukherjee and D.~Wichs, ``Two round multiparty computation via multi-key
  fhe,'' in \emph{Proceedings of the Annual International Conference on the
  Theory and Applications of Cryptographic Techniques}, Vienna, Austria, May
  2016, pp. 735--763.

\bibitem{li2017multi}
P.~Li, J.~Li, Z.~Huang, T.~Li, C.-Z. Gao, S.-M. Yiu, and K.~Chen, ``Multi-key
  privacy-preserving deep learning in cloud computing,'' \emph{Future
  Generation Computer Systems}, vol.~74, pp. 76--85, 2017.

\bibitem{wu2020efficient}
Y.~Wu, X.~Wang, W.~Susilo, G.~Yang, Z.~L. Jiang, Q.~Chen, and P.~Xu,
  ``Efficient server-aided secure two-party computation in heterogeneous mobile
  cloud computing,'' \emph{IEEE Transactions on Dependable and Secure
  Computing}, vol.~18, no.~6, pp. 2820--2834, 2021.

\bibitem{li2020npmml}
T.~Li, J.~Li, X.~Chen, Z.~Liu, W.~Lou, and T.~Hou, ``Npmml: A framework for
  non-interactive privacy-preserving multi-party machine learning,'' \emph{IEEE
  Transactions on Dependable and Secure Computing}, vol.~18, no.~4, pp.
  2969--2982, 2020.

\bibitem{bohli2013security}
J.-M. Bohli, N.~Gruschka, M.~Jensen, L.~L. Iacono, and N.~Marnau, ``Security
  and privacy-enhancing multicloud architectures,'' \emph{IEEE Transactions on
  Dependable and Secure Computing}, vol.~10, no.~4, pp. 212--224, 2013.

\bibitem{schiff2017screening}
G.~D. Schiff, L.~A. Volk, M.~Volodarskaya, D.~H. Williams, L.~Walsh, S.~G.
  Myers, D.~W. Bates, and R.~Rozenblum, ``Screening for medication errors using
  an outlier detection system,'' \emph{Journal of the American Medical
  Informatics Association}, vol.~24, no.~2, pp. 281--287, 2017.

\bibitem{gupta2018distributed}
O.~Gupta and R.~Raskar, ``Distributed learning of deep neural network over
  multiple agents,'' \emph{Journal of Network and Computer Applications}, vol.
  116, pp. 1--8, 2018.

\bibitem{vepakomma2018split}
P.~Vepakomma, O.~Gupta, T.~Swedish, and R.~Raskar, ``Split learning for health:
  Distributed deep learning without sharing raw patient data,'' \emph{arXiv
  preprint arXiv:1812.00564}, 2018.

\bibitem{ZLLSCC18}
Y.~Zhao, M.~Li, L.~Lai, N.~Suda, D.~Civin, and V.~Chandra, ``Federated learning
  with non-iid data,'' \emph{arXiv preprint arXiv:1806.00582}, 2018.

\bibitem{BEGHIIKKMM19}
K.~Bonawitz, H.~Eichner, W.~Grieskamp, D.~Huba, A.~Ingerman, V.~Ivanov,
  C.~Kiddon, J.~Kone{\v{c}}n{\`y}, S.~Mazzocchi, H.~B. McMahan \emph{et~al.},
  ``Towards federated learning at scale: System design,'' \emph{arXiv preprint
  arXiv:1902.01046}, 2019.

\bibitem{MSS19}
M.~Mohri, G.~Sivek, and A.~T. Suresh, ``Agnostic federated learning,'' in
  \emph{Proceedings of the International Conference on Machine Learning}, Long
  Beach, CA, June 2019, pp. 4615--4625.

\bibitem{bvhes20}
E.~Bagdasaryan, A.~Veit, Y.~Hua, D.~Estrin, and V.~Shmatikov, ``How to backdoor
  federated learning,'' in \emph{Proceedings of the International Conference on
  Artificial Intelligence and Statistics}, Sicily, Italy, Aug. 2020, pp.
  2938--2948.

\bibitem{fung2010privacy}
B.~C. Fung, K.~Wang, R.~Chen, and P.~S. Yu, ``Privacy-preserving data
  publishing: A survey of recent developments,'' \emph{ACM Computing Surveys},
  vol.~42, no.~4, pp. 1--53, 2010.

\bibitem{samarati1998protecting}
P.~Samarati and L.~Sweeney, ``Protecting privacy when disclosing information:
  k-anonymity and its enforcement through generalization and suppression,''
  Computer Science Laboratory, {SRI} International, Tech. Rep., 1998.

\bibitem{machanavajjhala2007diversity}
A.~Machanavajjhala, D.~Kifer, J.~Gehrke, and M.~Venkitasubramaniam,
  ``l-diversity: Privacy beyond k-anonymity,'' \emph{ACM Transactions on
  Knowledge Discovery from Data}, vol.~1, no.~1, pp. 24--24, 2007.

\bibitem{li2007t}
N.~Li, T.~Li, and S.~Venkatasubramanian, ``t-closeness: Privacy beyond
  k-anonymity and l-diversity,'' in \emph{Proceedings of the International
  Conference on Data Engineering}, Istanbul, Turkey, Apr. 2007, pp. 106--115.

\bibitem{aggarwal2008general}
C.~C. Aggarwal and S.~Y. Philip, ``A general survey of privacy-preserving data
  mining models and algorithms,'' in \emph{Privacy-Preserving Data Mining},
  2008, pp. 11--52.

\bibitem{ASR15}
Y.~A. A.~S. Aldeen, M.~Salleh, and M.~A. Razzaque, ``A comprehensive review on
  privacy preserving data mining,'' \emph{SpringerPlus}, vol.~4, no.~1, pp.
  1--36, 2015.

\bibitem{dwork2006differential}
C.~Dwork, ``Differential privacy,'' in \emph{Proceedings of the International
  Colloquium on Automata, Languages, and Programming}, Venice, Italy, July
  2006, pp. 1--12.

\bibitem{ACGMMTZ16}
M.~Abadi, A.~Chu, I.~Goodfellow, H.~B. McMahan, I.~Mironov, K.~Talwar, and
  L.~Zhang, ``Deep learning with differential privacy,'' in \emph{Proceedings
  of the ACM SIGSAC Conference on Computer and Communications Security},
  Vienna, Austria, Oct. 2016, pp. 308--318.

\bibitem{TF20}
A.~Triastcyn and B.~Faltings, ``Bayesian differential privacy for machine
  learning,'' in \emph{Proceedings of the International Conference on Machine
  Learning}, Shenzhen, China, July 2020, pp. 9583--9592.

\bibitem{ZHH20}
H.~Zheng, H.~Hu, and Z.~Han, ``Preserving user privacy for machine learning:
  local differential privacy or federated machine learning?'' \emph{IEEE
  Intelligent Systems}, vol.~35, no.~4, pp. 5--14, 2020.

\bibitem{DDR20}
J.~Dong, D.~Durfee, and R.~Rogers, ``Optimal differential privacy composition
  for exponential mechanisms,'' in \emph{Proceedings of the International
  Conference on Machine Learning}, Shenzhen, China, July 2020, pp. 2597--2606.

\bibitem{zhao2019privacy}
L.~Zhao, Q.~Wang, Q.~Zou, Y.~Zhang, and Y.~Chen, ``Privacy-preserving
  collaborative deep learning with unreliable participants,'' \emph{IEEE
  Transactions on Information Forensics and Security}, vol.~15, pp. 1486--1500,
  2019.

\bibitem{gursoy2019secure}
M.~E. Gursoy, A.~Tamersoy, S.~Truex, W.~Wei, and L.~Liu, ``Secure and
  utility-aware data collection with condensed local differential privacy,''
  \emph{IEEE Transactions on Dependable and Secure Computing}, vol.~18, no.~5,
  pp. 2365--2378, 2021.

\bibitem{gentry2009fully}
C.~Gentry, ``Fully homomorphic encryption using ideal lattices,'' in
  \emph{Proceedings of the Annual ACM Symposium on Theory of Computing},
  Bethesda, MD, June 2009, pp. 169--178.

\bibitem{yao1982protocols}
A.~C. Yao, ``Protocols for secure computations,'' in \emph{Proceedings of the
  Annual Symposium on Foundations of Computer Science}, Washington, DC, Nov.
  1982, pp. 160--164.

\bibitem{graepel2012ml}
T.~Graepel, K.~Lauter, and M.~Naehrig, ``Ml confidential: Machine learning on
  encrypted data,'' in \emph{Proceedings of the International Conference on
  Information Security and Cryptology}, Seoul, Korea, Nov. 2012, pp. 1--21.

\bibitem{hesamifard2017cryptodl}
E.~Hesamifard, H.~Takabi, and M.~Ghasemi, ``Cryptodl: Deep neural networks over
  encrypted data,'' \emph{arXiv preprint arXiv:1711.05189}, 2017.

\bibitem{yu2013toward}
J.~Yu, P.~Lu, Y.~Zhu, G.~Xue, and M.~Li, ``Toward secure multikeyword top-k
  retrieval over encrypted cloud data,'' \emph{IEEE Transactions on Dependable
  and Secure Computing}, vol.~10, no.~4, pp. 239--250, 2013.

\bibitem{wagh2019securenn}
S.~Wagh, D.~Gupta, and N.~Chandran, ``Secure{NN}: 3-party secure computation
  for neural network training,'' \emph{Proceedings of the Privacy Enhancing
  Technologies}, vol. 2019, no.~3, pp. 26--49, 2019.

\bibitem{mohassel2017secureml}
P.~Mohassel and Y.~Zhang, ``Secureml: A system for scalable privacy-preserving
  machine learning,'' in \emph{Proceedings of the IEEE Symposium on Security
  and Privacy}, San Jose, CA, May 2017, pp. 19--38.

\bibitem{juvekar2018gazelle}
C.~Juvekar, V.~Vaikuntanathan, and A.~Chandrakasan, ``{GAZELLE}: A low latency
  framework for secure neural network inference,'' in \emph{Proceedings of the
  USENIX Security Symposium}, Santa Clara, CA, Aug. 2018, pp. 1651--1669.

\bibitem{liu2017oblivious}
J.~Liu, M.~Juuti, Y.~Lu, and N.~Asokan, ``Oblivious neural network predictions
  via minionn transformations,'' in \emph{Proceedings of the ACM SIGSAC
  Conference on Computer and Communications Security}, Dallas, TX, Nov. 2017,
  pp. 619--631.

\bibitem{RRDNA22}
A.~Resende, D.~Railsback, R.~Dowsley, A.~C. Nascimento, and D.~F. Aranha,
  ``Fast privacy-preserving text classification based on secure multiparty
  computation,'' \emph{IEEE Transactions on Information Forensics and
  Security}, vol.~17, pp. 428--442, 2022.

\bibitem{XMSSXJ18}
Z.~Xia, X.~Ma, Z.~Shen, X.~Sun, N.~N. Xiong, and B.~Jeon, ``Secure image lbp
  feature extraction in cloud-based smart campus,'' \emph{IEEE Access}, vol.~6,
  pp. 30\,392--30\,401, 2018.

\bibitem{ball2016garbling}
M.~Ball, T.~Malkin, and M.~Rosulek, ``Garbling gadgets for boolean and
  arithmetic circuits,'' in \emph{Proceedings of the ACM SIGSAC Conference on
  Computer and Communications Security}, Vienna, Austria, October 2016, pp.
  565--577.

\bibitem{boyle2018limits}
E.~Boyle, Y.~Ishai, and A.~Polychroniadou, ``Limits of practical sublinear
  secure computation,'' in \emph{Proceedings of the Annual International
  Cryptology Conference}, Santa Barbara, CA, August 2018, pp. 302--332.

\bibitem{mohassel2018aby3}
P.~Mohassel and P.~Rindal, ``Aby3: A mixed protocol framework for machine
  learning,'' in \emph{Proceedings of the ACM SIGSAC Conference on Computer and
  Communications Security}, New York, NY, October 2018, pp. 35--52.

\bibitem{imani2019framework}
M.~Imani, Y.~Kim, S.~Riazi, J.~Messerly, P.~Liu, F.~Koushanfar, and T.~Rosing,
  ``A framework for collaborative learning in secure high-dimensional space,''
  in \emph{Proceedings of the IEEE International Conference on Cloud
  Computing}, Milan, Italy, July 2019, pp. 435--446.

\bibitem{ryffel2018generic}
T.~Ryffel, A.~Trask, M.~Dahl, B.~Wagner, J.~Mancuso, D.~Rueckert, and
  J.~Passerat-Palmbach, ``A generic framework for privacy preserving deep
  learning,'' \emph{arXiv preprint arXiv:1811.04017}, 2018.

\bibitem{kamara2014scaling}
S.~Kamara, P.~Mohassel, M.~Raykova, and S.~Sadeghian, ``Scaling private set
  intersection to billion-element sets,'' in \emph{Proceedings of the
  International Conference on Financial Cryptography and Data Security}.\hskip
  1em plus 0.5em minus 0.4em\relax Christ Church, Barbados: Springer, March
  2014, pp. 195--215.

\bibitem{carter2016secure}
H.~Carter, B.~Mood, P.~Traynor, and K.~Butler, ``Secure outsourced garbled
  circuit evaluation for mobile devices,'' \emph{Journal of Computer Security},
  vol.~24, no.~2, pp. 137--180, 2016.

\bibitem{yao1986generate}
A.~C.-C. Yao, ``How to generate and exchange secrets,'' in \emph{Proceedings of
  the Annual Symposium on Foundations of Computer Science}, Washington, DC,
  Oct. 1986, pp. 162--167.

\bibitem{shamir1979share}
A.~Shamir, ``How to share a secret,'' \emph{Communications of the ACM},
  vol.~22, no.~11, pp. 612--613, 1979.

\bibitem{kamara2011outsourcing}
S.~Kamara, P.~Mohassel, and M.~Raykova, ``Outsourcing multi-party
  computation.'' \emph{IACR Cryptol. Eprint Arch.}, vol. 2011, p. 272, 2011.

\bibitem{nikolaenko2013privacy}
V.~Nikolaenko, U.~Weinsberg, S.~Ioannidis, M.~Joye, D.~Boneh, and N.~Taft,
  ``Privacy-preserving ridge regression on hundreds of millions of records,''
  in \emph{Proceedings of the IEEE Symposium on Security and Privacy}, San
  Francisco, CA, May 2013, pp. 334--348.

\bibitem{evans2017pragmatic}
D.~Evans, V.~Kolesnikov, and M.~Rosulek, ``A pragmatic introduction to secure
  multi-party computation,'' \emph{Foundations and Trends{\textregistered} in
  Privacy and Security}, vol.~2, no. 2-3, 2017.

\bibitem{yang2019federated}
Q.~Yang, Y.~Liu, Y.~Cheng, Y.~Kang, T.~Chen, and H.~Yu, ``Federated learning,''
  \emph{Synthesis Lectures on Artificial Intelligence and Machine Learning},
  vol.~13, no.~3, pp. 1--207, 2019.

\bibitem{DTSKHR18}
S.~A. Osia, A.~Taheri, A.~S. Shamsabadi, K.~Katevas, H.~Haddadi, and H.~R.
  Rabiee, ``Deep private-feature extraction,'' \emph{IEEE Transactions on
  Knowledge and Data Engineering}, vol.~32, no.~1, pp. 54--66, 2018.

\bibitem{goodfellow2016deep}
I.~Goodfellow, Y.~Bengio, and A.~Courville, \emph{Deep learning}.\hskip 1em
  plus 0.5em minus 0.4em\relax MIT press, 2016.

\bibitem{lecun1998gradient}
Y.~LeCun, L.~Bottou, Y.~Bengio, and P.~Haffner, ``Gradient-based learning
  applied to document recognition,'' \emph{Proceedings of the IEEE}, vol.~86,
  no.~11, pp. 2278--2324, 1998.

\bibitem{shokri2017membership}
R.~Shokri, M.~Stronati, C.~Song, and V.~Shmatikov, ``Membership inference
  attacks against machine learning models,'' in \emph{Proceedings of the IEEE
  Symposium on Security and Privacy}, Santa Clara, CA, Aug. 2017, pp. 3--18.

\bibitem{ding2022privacy}
X.~Ding, H.~Fang, Z.~Zhang, K.-K.~R. Choo, and H.~Jin, ``Privacy-preserving
  feature extraction via adversarial training,'' \emph{IEEE Transactions on
  Knowledge and Data Engineering}, vol.~34, no.~4, pp. 1967--1979, 2022.

\bibitem{cai2022privacy}
G.~Cai, X.~Wei, and Y.~Li, ``Privacy-preserving cnn feature extraction and
  retrieval over medical images,'' \emph{International Journal of Intelligent
  Systems}, 2022.

\end{thebibliography}

\begin{IEEEbiography}[{\includegraphics[width=1in,height=1.25in,clip,keepaspectratio]{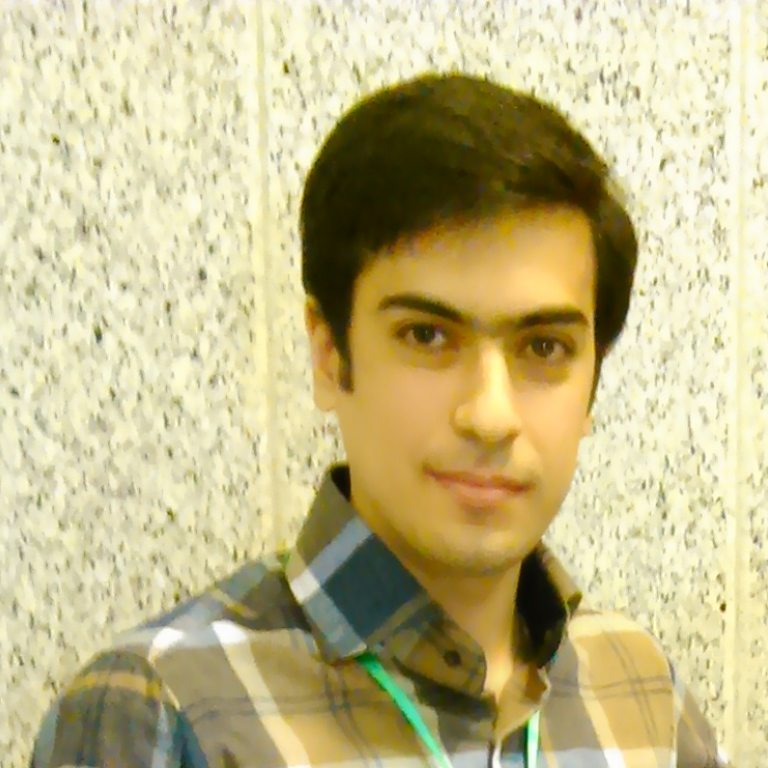}}]{Alireza Sarmadi}
received his B.Sc. degree in electrical engineering from the University of Tehran, Tehran, Iran in 2016. He is a Ph.D. candidate at New York University (NYU) and a Research Assistant at the Control/Robotics Research Laboratory, at NYU. His research interests include robotics, control systems, machine learning, and biomechanics.
\end{IEEEbiography}

\begin{IEEEbiography}[{\includegraphics[width=1in,height=1.25in,clip,keepaspectratio]{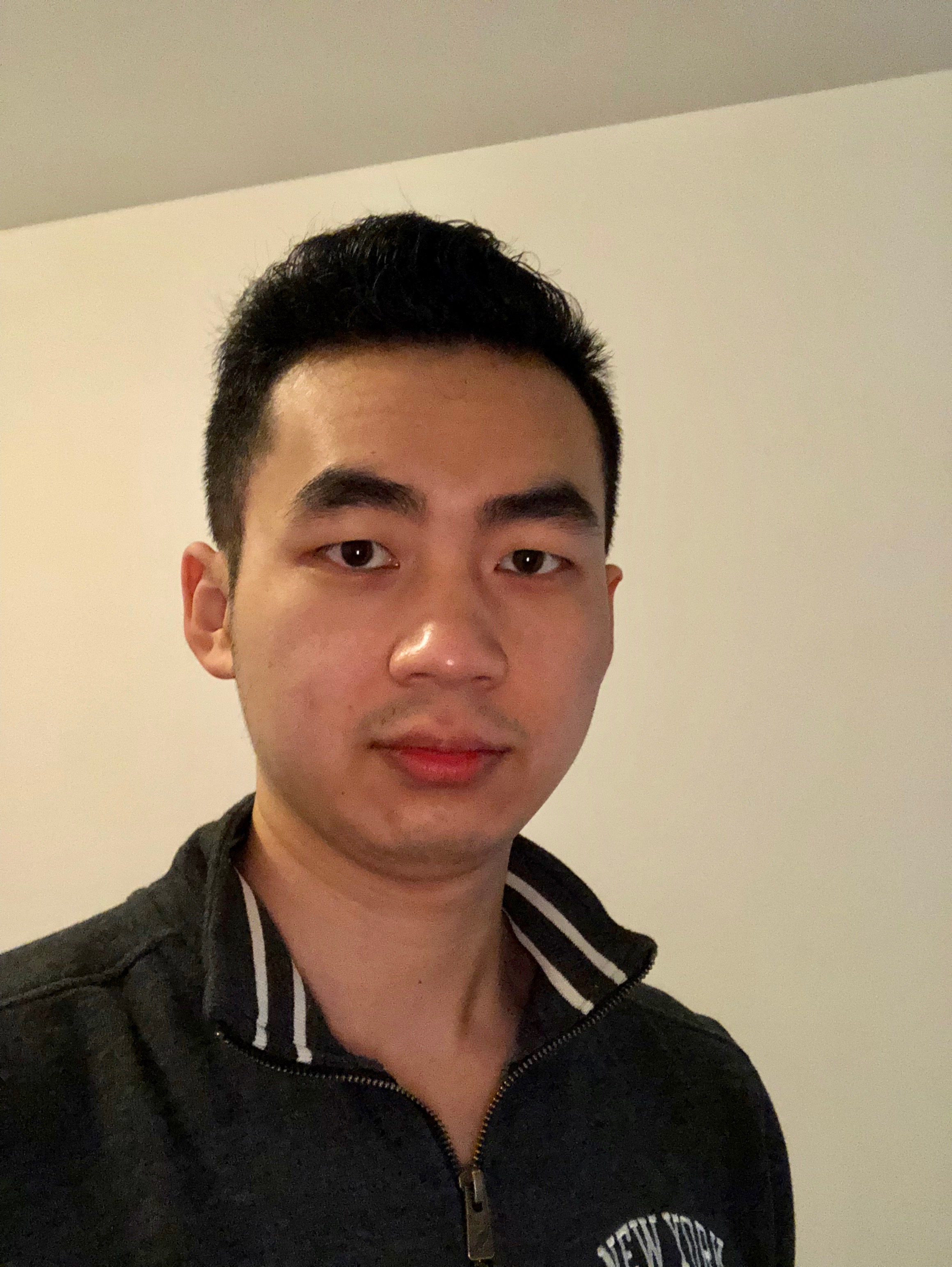}}]{Hao Fu}
was born on November 5, 1994, in Anyang, China. He is a Ph.D. candidate in the Department of Electrical and Computer Engineering at New York University, Tandon School of Engineering, Brooklyn, NY, USA. In 2019, he received his Master of Science degree in Electrical Engineering in the same department as well. He received his Bachelor of Science degree in Physics from the University of Science and Technology of China, Hefei, China, in 2017. His major field of study contains machine learning, finance, and control theory. From 2017 to 2018, he was a research assistant in the NYU Wireless lab. In 2018 Fall, he joined in Control/Robotics Research Laboratory (CRRL). Previously, he was studying the possibility of using machine learning tools to develop economical navigation algorithms. Additionally, he was also studying the possibility of using neural networks to assist decision-making in finance. Currently, he is studying backdooring attacks against neural networks and security problems in cyber-physical systems. He published one article at IEEE Conference on Control Technology and Applications in 2020. 

\end{IEEEbiography}

\begin{IEEEbiography}[{\includegraphics[width=1in,height=1.25in,clip,keepaspectratio]{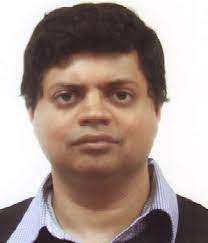}}]{Prashanth Krishnamurthy}
received the B.Tech. degree in electrical engineering from the Indian Institute of Technology Madras, Chennai, in 1999, and the M.S. and Ph.D. degrees in electrical engineering from Polytechnic University (now NYU), Brooklyn, NY, USA, in 2002 and 2006, respectively. He is currently a Research Scientist with the Department of Electrical and Computer Engineering, NYU Tandon School of Engineering, New York, and a Senior Researcher with FarCo Technologies. He has co-authored over 130 journal articles and conference papers in the broad areas of autonomous systems, robotics, and control systems. He has also co-authored the book Modeling and Adaptive Nonlinear Control of Electric Motors (Springer-Verlag, 2003). His research interests include autonomous vehicles and robotic systems, multi-agent systems, sensor data fusion, robust and adaptive nonlinear control, resilient control, machine learning, real-time embedded systems, cyber-physical systems and cyber-security, decentralized and large-scale systems, and real-time software implementations.
\end{IEEEbiography}

\begin{IEEEbiography}[{\includegraphics[width=1in,height=1.25in,clip,keepaspectratio]{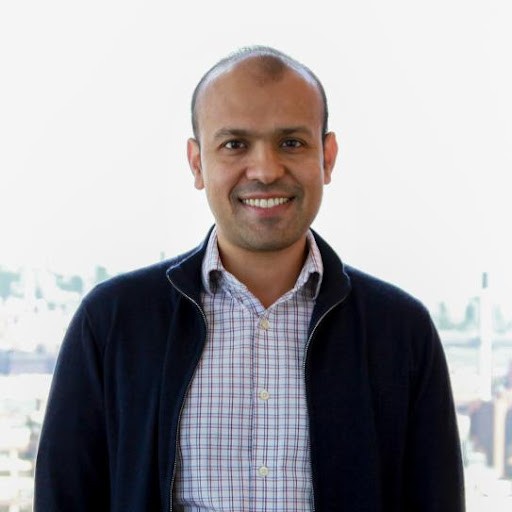}}]{Siddharth Garg}
received the B.Tech. degree in electrical engineering from the Indian Institute of Technology Madras, Chennai, India, and the Ph.D. degree in electrical and computer engineering from Carnegie Mellon University, Pittsburgh, PA, in 2009. He is currently an Associate Professor at New York University, New York, where he joined as an Assistant Professor in 2014. Prior to this, he was an Assistant Professor with the University of Waterloo, Waterloo, ON, Canada, from 2010 to 2014. His current research interests include computer engineering, and more particularly in secure, reliable, and energy efficient computing. He was a recipient of the NSF Career Award in 2015, and the paper awards at the IEEE Symposium on Security and Privacy 2016, the USENIX Security Symposium in 2013, the Semiconductor Research Consortium TECHCON in 2010, and the International Symposium on Quality in Electronic Design in 2009. He was listed in popular science magazine’s annual list of “Brilliant 10” researchers. He serves on the technical program committee of several top conferences in the area of computer engineering and computer hardware and has served as a reviewer for several IEEE and ACM journals conferences.
\end{IEEEbiography}

\begin{IEEEbiography}[{\includegraphics[width=1in,height=1.25in,clip,keepaspectratio]{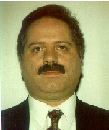}}]{Farshad Khorrami}
received the bachelor’s degrees in mathematics and in electrical engineering from The Ohio State University in 1982 and 1984, respectively, and the master’s degree in mathematics and the Ph.D. degree in electrical engineering from The Ohio State University, in 1984 and 1988, respectively. He is currently a Professor with the Electrical and Computer Engineering Department, NYU, where he joined as an Assistant
Professor in September 1988. He has developed and directed the Control/Robotics Research Laboratory, Polytechnic University (Now NYU). His research has been supported by the DARPA, ARO, NSF, ONR, AFRL, Sandia National Laboratory, ARL, NASA, and Corps. He has published more than 300 refereed journal articles and conference papers in these areas. His book on Modeling and Adaptive Nonlinear Control of Electric Motors (Springer-Verlag, 2003). He also holds 14 U.S. patents on novel smart micropositioners and actuators, control systems, cyber security, and wireless sensors and actuators. His research interests include adaptive and nonlinear controls, robotics and automation, unmanned vehicles (fixed-wing and rotary wing aircrafts as well as underwater vehicles and surface ships), resilient control and cyber security for cyber-physical systems, large-scale systems, decentralized control, and real-time embedded instrumentation and control. He has served as the general chair and a conference organizing committee member for several international
conferences.
\end{IEEEbiography}
\end{document}